\pdfoutput=1

\documentclass[10pt, a4paper]{article}
\usepackage{lrec-coling2024}
\usepackage{graphicx}
\usepackage{times}
\usepackage{amsmath}
\usepackage{latexsym}
\usepackage[T1]{fontenc}
\usepackage{float}
\usepackage[utf8]{inputenc}
\usepackage{microtype}
\usepackage{multirow}
\usepackage{mdframed}
\usepackage{graphicx} 
\usepackage{caption}

\title{Using Persuasive Writing Strategies to Explain and Detect Health Misinformation}

\name{Danial Kamali\textsuperscript{*}, Joseph Romain\textsuperscript{*}, Huiyi Liu\textsuperscript{\dag}, Wei Peng\textsuperscript{\ddag}, Jingbo Meng\textsuperscript{\S},
\\ {\bf \large Parisa Kordjamshidi\textsuperscript{*}}  }

\address{\textsuperscript{*}Department of Computer Science and Engineering, Michigan State University,\\ \textsuperscript{\dag}Department of Communication, Michigan State University,\\ 
\textsuperscript{\ddag}Department of Media and Information, Michigan State University,\\ 
\textsuperscript{\S}School of Communication, Ohio State University \\
\{kamalida,romainj2,liuhui5,pengwei\}@msu.edu, meng.28@osu.edu, kordjams@msu.edu}

\abstract{Nowadays, the spread of misinformation is a prominent problem in society. Our research focuses on aiding the automatic identification of misinformation by analyzing the persuasive strategies employed in textual documents. We introduce a novel annotation scheme encompassing common persuasive writing tactics to achieve our objective. Additionally, we provide a dataset on health misinformation, thoroughly annotated by experts utilizing our proposed scheme. 
Our contribution includes proposing a new task of annotating pieces of text with their persuasive writing strategy types. We evaluate fine-tuning and prompt-engineering techniques with pre-trained language models of the BERT family and the generative large language models of the GPT family using persuasive strategies as an additional source of information.
We evaluate the effects of employing persuasive strategies as intermediate labels in the context of misinformation detection. Our results show that those strategies enhance accuracy and improve the explainability of misinformation detection models. The persuasive strategies can serve as valuable insights and explanations, enabling other models or even humans to make more informed decisions regarding the trustworthiness of the information.
\\ \newline \Keywords{Misinformation Detection, Explainable AI, Persuasive Writing Strategy, Large Language Model}}

\begin{document}

\maketitleabstract

\section{Introduction}

The concept of misinformation has long been present, but in recent years, there has been an alarming surge in the dissemination of inaccurate information, particularly concerning health and the medical industry~\cite{info:doi/10.2196/17187}. This influx of misinformation has had detrimental effects on society as a whole. People's trust in news organizations has fallen drastically, resulting in a less informed populace~\cite{islam2020covid}. Many news resources purposefully make false news articles to spread misinformation and increase hostility between political groups. Most prior online misinformation research focuses on the political context~\cite{lazer2018science}. However, health-related misinformation has grown exponentially during the COVID-19 pandemic, and these types of misinformation have a significant impact on individuals and society~\cite{loomba2021measuring}. There are vast amounts of data for human fact-checkers, more than what they can go through manually, so professionals have turned to machine learning for automated misinformation identification for years, which is still a prominent challenge.

\begin{figure*}
\centering
  \includegraphics[width=0.96\textwidth]{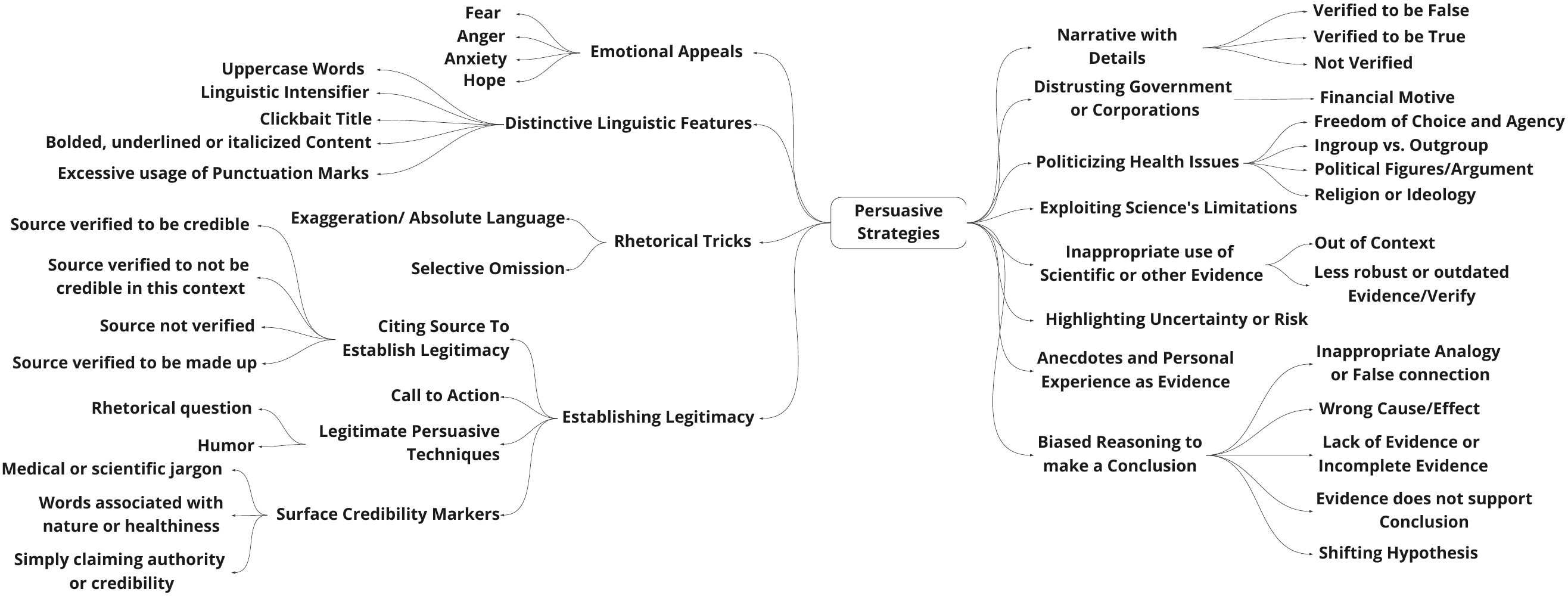}
 \caption{Persuasive Strategies annotation hierarchy scheme.}
    \label{labels-tree}

\end{figure*}

For people to trust a machine's verdict in misinformation, the models must produce interpretable reasoning as to why an article is incorrect. This paper presents a novel annotation scheme containing a comprehensive hierarchy of persuasive writing strategies systematically identified from online misinformation~\cite{doi:10.1080/1369118X.2022.2085615}, as shown in Figure~\ref{labels-tree}. We train models to spot pieces of an article that can potentially make it untrustworthy. By understanding these tactics in the information source, users can have a deeper analysis and are less likely to be misled by false information.
Our contributions include 1) A new corpus annotated with comprehensive hierarchical persuasive writing strategy labels. 2) A new challenging task of characterizing the persuasive writing strategies in text and developing baselines. 3) Demonstrating the integration of persuasive strategies enhances misinformation detection accuracy and offers insightful explanations that elucidate the underlying reasons for classifying particular information as false.

\section{Related Work}

Systematic research on rumors in mass media and deception in interpersonal communication dates back more than 70 years ago~\cite{allport1945wartime,knapp1944psychology}. Nowadays, the significance of misinformation is increasingly evident as it effortlessly transcends temporal and geographical boundaries. Furthermore, the ease of searching, archiving, and accessing the information~\cite{fernandez2015health} exacerbates the situation~\cite{vosoughi2018spread,shu2017fake}. 

In this study, we focus on health misinformation ~\cite{doi:10.1080/1369118X.2022.2085615,swire2019public}. Before the COVID-19 pandemic, health misinformation had already attracted the attention of researchers due to the insurgence of childhood vaccination misinformation on social media~\cite{wang2019systematic}. However, the burgeoning health misinformation during the COVID-19 pandemic \cite{kouzy2020coronavirus,cuan2020misinformation} and its associated negative impacts (e.g., vaccination hesitancy~\cite{loomba2021measuring, roozenbeek2020susceptibility} has brought heightened attention from the research community to this pressing issue.
  

With the vast daily news output, it is unrealistic for human fact-checkers to verify every detail. As a result, many have turned to machine learning models to aid the process of claim verification and misinformation detection in its diverse forms~\cite{https://doi.org/10.1002/cpe.6479,yuliani2019hoax,8468301, burfoot2009automatic,de-sarkar-etal-2018-attending,https://doi.org/10.48550/arxiv.2007.08024, khanday2021identifying}. 
Here, we overview the datasets, models, and techniques that are specifically relevant to health misinformation. 

\noindent\textbf{Datasets.} Many datasets are constructed for training and evaluating misinformation detection models. These datasets tend to vary significantly in attributes and content. All datasets contain claims and labels from fact-checking websites such as Politifact~\cite{DBLP:journals/corr/Wang17j}, Snopes~\cite{10.1145/2983323.2983661}, or both~\cite{10.1145/3331184.3331248}. In recent years, more datasets have been created by crawling various fact-checking websites to increase generalizability~\cite{DBLP:journals/corr/abs-2006-11343}. Besides claims and labels, datasets often include evidence pages and metadata from sources such as Google~\cite{augenstein-etal-2019-multifc} and Wikipedia~\cite{https://doi.org/10.48550/arxiv.2106.05707,yang-etal-2022-coarse}. Recently, \citet{yang-etal-2022-coarse} constructed two explainable fake news datasets to help develop explainable models. Our dataset uses a subset of claims from the MultiFC dataset~\cite{augenstein-etal-2019-multifc} that we augmented with supporting articles for our in-depth context analysis purposes.

\noindent\textbf{Existing Models.} 
Different approaches have been adopted to address automated misinformation detection. One research branch utilizes Natural Language Processing (NLP) techniques and large language models (LLMs) to detect misinformation within news texts \cite{manzoor2019fake,zhang2023llmbased,jiang2023active}. Recent studies highlight the beneficial impact of incorporating syntactic information into transformer models, as demonstrated by ~\citet{kamali2023syntax}, and its specific advantages in detecting misinformation ~\cite{10.1371/journal.pone.0256940}. Another emerging research area focuses on identifying misinformation by examining specific textual features. For instance, \citet{ghanem2019emotional} highlighted that falsified information often carries unique emotional signatures. Building on this observation, \citet{DBLP:journals/corr/abs-2101-09810} introduced FakeFlow, a model that captures the flow of affective information and emotions within an article. This can assist in identifying articles that strategically manipulate reader emotions. Furthermore, the DEF~\cite{LUVEMBE2023103354} study  considers both the emotion of the publisher and the social emotions it provokes, potentially broadening the context for detecting misinformation.
Recent studies have increasingly underscored the utility of LLMs. Notably, \citet{lee-etal-2020-language} employed LLMs such as GPT-2~\cite{Radford2019LanguageMA} to assess claim factuality through evidence-conditioned perplexity. While there remains a scarcity of research on LLM misinformation detection~\cite{zhang2023llmbased}, emerging works~\cite{press2023measuring,pelrine2023reliable,yao2023react,jiang2023active} suggest that fusing LLM reasoning with external knowledge can substantially bolster performance in reasoning-intensive NLP tasks.
Compared to approaches like FakeFlow, which centers on emotional manipulation, our research adopts a more holistic perspective. We investigate a broader spectrum of factors through the lens of persuasive strategies. This approach not only enhances our ability to detect misinformation but also improves the interpretability framework. Consequently, this framework further aids non-communication experts in misinformation detection \cite{doi:10.1177/20539517211013868}. In contrast to LLM-based approaches such as those by \citet{jiang2023active,zhang2023llmbased}, our model labels misinformation without requiring access to any external knowledge base.

\section{Persuasive Writing Strategies}
{Persuasion is an essential part of news misinformation. Misinformation often plays on people's suspicions, premises, and biases to create narratives that feel plausible and compelling~\cite{vosoughi2018spread}. By employing common persuasive techniques such as appeals to emotion, anecdotal evidence, and bandwagon appeals, misinformation can seem more credible than it actually is. Previous studies \cite{eisend2022persuasion, guess2020digital} have demonstrated that when individuals become aware of the persuasive tactics or have adequate media or information literacy to understand these persuasive attempts, they become less susceptible to persuasion attempts such as political propaganda and advertising. Such media literacy interventions coupled with fact-checking were found to be more effective than each component used alone~\cite{hameleers2022separating}. 
}

\subsection{Persuasive Strategies Serve as Explanations}

The deliverable of the current misinformation models usually indicates whether the information is true or false or evidence supporting or denying claims. A single label can only have a limited effect in diminishing the influence of misinformation~\cite{margolin2018political}. One step forward from the simple fact-checking label is to provide an explainable justification for why specific claims are misleading~\cite{kotonya2020explainable, atanasova2020generating}. The research in communications has long discovered that providing justification and understanding the underlying techniques used in claims to influence the information receiver will be critical to helping individuals cope with persuasive attempts \cite{friestad1994persuasion}. The current study attempts to automatically annotate persuasive features in misinformation, which can serve as intermediate features for classifying misinformation and as auxiliary information for future media literacy training tools to assist individuals in better misinformation detection.   

\subsection{Persuasive Strategies in Health Misinformation}

The persuasive features implemented in the current study were based on a review that systematically screened 1,700 research articles related to online health misinformation to identify 12 thematic groups of persuasive strategies from 58 eligible articles~\cite{doi:10.1080/1369118X.2022.2085615}. For instance, for the theme of politicizing health issues, there are subgroups such as the trope of freedom and choice, the rhetoric of ingroup vs. outgroup, citing political figures or political arguments, and the use of religion and ideology. Although some similar approaches are available, such as bias detection based on political ideology~\cite{baly-etal-2020-detect}, propaganda detection based on linguistic features~\cite{rashkin-etal-2017-truth} or using multiple categories of features~\cite{da-san-martino-etal-2019-fine, da-san-martino-etal-2020-prta, dimitrov-etal-2021-semeval}, persuasive strategies based on a particular theory such as Aristotle's rhetoric theory~\cite{chen2021persuasion} or Moral Foundation Theory~\cite{lin2018acquiring}, the persuasive writing techniques identified and annotated in our study have a competitive advantage because they were based on a systematic and comprehensive review of current literature to provide a full-scale framework.

\section{Dataset Construction}
\label{section:dataset}

\subsection{Health-related Subset from MultiFC}

To build our persuasive writing strategy dataset, we have chosen to use a subset of the MultiFC~\cite{augenstein-etal-2019-multifc} dataset, consisting of claims labeled as health, medical, or food, containing 599 claims (7 duplicates). We refer to this as \textit{Health subset}. {Each sample of the MultiFC dataset includes a claim (the main point of a news article), a few sentences of supporting articles with their links, and a domain label providing the context for each claim.}  The misinformation labels for this subset are \textit{mostly true}, \textit{truth!}, \textit{true}, \textit{disputed!}, \textit{in the green}, \textit{in the red}, \textit{mostly truth!}, \textit{mostly false}, \textit{fiction!}, \textit{false}. The \textit{unproven} (215 claims), \textit{satirical} (22 claims), and non-health or non-medical related articles (33 claims) were excluded or filtered out through an expert review process, leaving a total of 329 entries.
 
\subsection{Retrieval of Full Articles for Annotation}

 We manually retrieved the complete article of each claim from the MultiFC health subset by searching the claim on the Google search engine because the URLs provided in the MultiFC mostly turned invalid. A health communication master's student used the claims or the keywords in the claim to conduct the search and browse the search results to identify the full article that matched the claim. Out of the 329 claims investigated, 86 original articles were either deleted from the internet or existed in a format other than text, resulting in a total of \textbf{242} articles containing \textbf{5666} sentences. Three expert annotators annotated the remaining articles with persuasive writing labels using the WebAnno annotation tool \cite{eckart-de-castilho-etal-2016-web}.

\subsection{Annotation of Full Articles}

\subsubsection{Annotation Codebook Development}
Our annotation scheme organizes the persuasive writing strategies into three hierarchy levels, as shown in Figure \ref{labels-tree}. We call the levels the parent tag level, the child tag level, and the sub-tag level. The parent tag level is the overarching group of strategies, corresponding to 12 thematic groups of persuasive strategies identified in \citet{doi:10.1080/1369118X.2022.2085615}, such as \textit{Politicizing Health Issues}. The child tag level is the subgroup of more specific strategies, e.g., \textit{Politicizing Health Issues- Religion/Ideology}, \textit{Emotional Appeals-Fear}. Although the original framework of \citet{doi:10.1080/1369118X.2022.2085615} had only two levels, in the codebook development process, based on health communication and persuasion theories, a third level, i.e., sub-tag level, was added by the health communication experts in our team to provide even more specificity when needed, e.g., \textit{Establishing legitimacy- Surface Credibility Markers- Medical/Scientific jargon}. These additional subgroup child tags and the sub-tag level were added in our paper to the original framework in~\citet{doi:10.1080/1369118X.2022.2085615}.

\begin{figure}
    \centering
  \includegraphics[width=0.48\textwidth,]{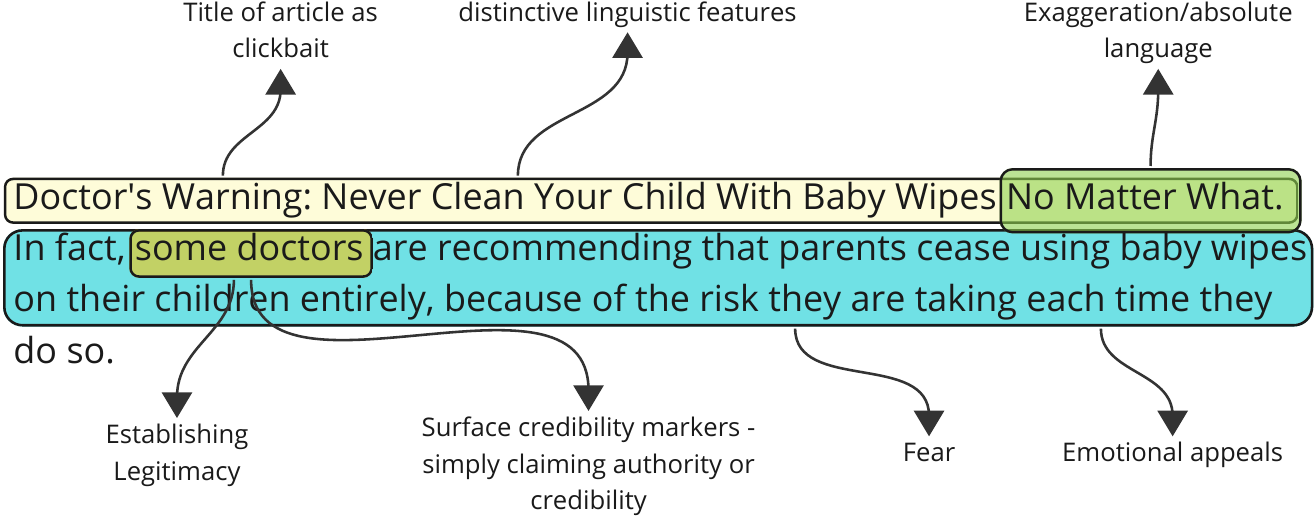}
 \caption{Persuasive strategy labeling example.}
\label{ps:example}
\end{figure}

In our labeling scheme, we impose a hierarchical structure; this means that text can be annotated with a child tag without being annotated with any of its sub-tags. However, if a text sequence is annotated with a child tag, it must be annotated with its respective parent tag. If a text sequence is annotated with a sub-tag, it must also be annotated with its respective child tag. Figure \ref{ps:example} shows an example of persuasive strategy labeling. 
Although the annotations can apply to any piece of text and any level of granularity, such as phrases or sentences, as we will see in the task specification, we assign the strategies to the sentences containing them.

Two communication professors and a master's student specializing in health and risk communication, all our team members, developed and refined an annotation codebook for these three-level labels.
 The master's student annotated all the articles. Note that the annotators only annotated the persuasive strategies. The claim of each article was assumed to be the central idea of the full article. The label or annotation of the claim was imported from MultiFC or their origin in MultiFC, which used fact-checking information to assign the label. In other words, we did not need to deal with the bias among the annotators regarding judgment on whether the claim or the whole article included false information.

\subsubsection{Inter Annotator Agreement}

A random sample of 25 articles (10 percent of the entire sample) was coded by the master's student and the two professors. The inter-annotator reliability was satisfactory with the following details: The average pairwise percentage agreement was 88.99\%; pairwise agreement between the master's student and the two professors was 88.77\% and 86.78\%, respectively; and pairwise agreement between the two professors was 91.41\%. The average pairwise Cohen's Kappa was 0.88; pairwise Cohen's Kappa between the master's student and the two professors were 0.88 and 0.86, respectively; and pairwise Cohen's Kappa between the two professors was 0.91. The master's student annotated all the articles and then double-checked by at least one of the professors to resolve any discrepancies.
The detailed frequency of each persuasive strategy can be found in Appendix \ref{appendix:data}.

\section{Approach}

We follow three learning tasks in this paper.\\
\noindent\textbf{Task 1 (Misinformation Detection):} Given a claim $C$ and an article $A$, a model should classify the article with its trustworthiness label, $Tw$. In the simplest case, $Tw$ can be a binary label indicating the document's truth. However, detecting the veracity of text is often more complex; thus, these label sets typically include more values, such as \textit{Mixed}, \textit{Unverifiable}, and \textit{Outdated}.

\noindent\textbf{Task 2 (Persuasive Strategy Labeling):} Given an article $A$ represented as a sequence of $n$ sentences, $A=\{s_1, s_2, ...s_n\}$, the task is to assign each sentence $s_i$ a set of persuasive strategy labels in $L=\{l_{1}, l_{2}, ...l_{m}\}$. A boolean value $l_{ij}$ indicates whether sentence $s_i$ follows persuasive strategy $l_j$. Each sentence $s_i$ can carry multiple labels, indicating the presence of several strategies. In other words, sentences could be marked with multiple labels in the hierarchy of persuasive writing strategies or none of them. 
In our setting, $L$ follows a label hierarchy shown in Figure~\ref{labels-tree}. 
In this task, the surrounding context can be used as a part of the input information for each sentence classification. The input sentence can be evaluated with three levels of context: \textit{None}, \textit{Low}, and \textit{High}. The \textit{Low} context includes two additional surrounding sentences, one to the left of the target sentence and one to the right. \textit{High} context includes four additional sentences, two to the left of the target sentence and two to the right.
While this task can be performed at the sentence or paragraph level, we found the sentence-level assignment of the labels to be more effective.

\noindent\textbf{Task 3 (Combining Persuasive Strategies \& Misinformation Detection): }
Since the persuasive strategies potentially can help both the detection and characterization of the news document, we define this third combined task. 
The input includes the claim $C$ and article $A$. The output is the sentence-level persuasive strategy labels $l_{ij}$ and the misinformation label $Tw$. This task can be modeled as a pipeline or a joint labeling setting.

\subsection{Models}
\label{subsection:models}

 In recent years, foundation models have emerged as prominent tools, excelling in numerous NLP problems and establishing themselves as robust baselines~\cite{foundation}. In our experiments, we employ two distinct types of foundation models. Firstly, we utilize RoBERTa~\cite{liu2019roberta}, a pre-trained encoder-based model extensively adopted across the NLP community, trained using masked language modeling. Furthermore, we use GPT-3~\cite{gpt} and the more recent GPT-4~\cite{openai2023gpt4}, both of which have significantly larger parameter configurations, serving as our generative language models.
For all tasks and experiments, we fine-tune RoBERTa-based models. We pass the RoBERTa model's CLS token result to a classification layer to perform classification tasks. GPT-3 and GPT-4 are used in the zero-shot setting due to their high training cost.

\section{Experiments}
The goal of our experiments and analysis is to answer the following research questions: 
\medskip

\noindent Q1. Do the foundation models provide reasonable accuracy on the misinformation detection task compared to earlier results? (Section \ref{section:baseline-comparison}) 
\medskip

\noindent Q2. How difficult is it to predict the various persuasive writing strategies in sentences after fine-tuning the baseline foundation models? (Section \ref{section:persuasive})
\medskip

\noindent Q3. How well do the foundation models predict the article's veracity in \textit{Health subset}? (Section. \ref{section:misinfor-with-persuasive})
\medskip

\noindent Q4. How will including persuasive strategy labels affect a model's ability to detect misinformation in the articles? (Section \ref{section:misinfor-with-persuasive})
\medskip

\noindent\textbf{Experimental details.} We ran all experiments using PyTorch~\cite{pyTorch}. We used the pandas~\cite{pandas} library for data manipulation and the NLTK~\cite{nltk} and spacy~\cite{spacy} libraries for tokenization. Our
reported results are the average of three runs with
different random seeds on an NVIDIA A6000 GPU. The details of the hyperparameters of our experiments are in Appendix \ref{appendix:hyper}. Our code, models, and data are publicly available\footnote{\url{https://github.com/HLR/Misinformation-Detection}}.

\subsection{Competency of Foundation Models}
\label{section:baseline-comparison}
In this section, we demonstrate a small experiment to evaluate the effectiveness of our baseline foundation model in misinformation detection to answer Question 1. For a broader view of the evaluation of a variety of language models, we refer the reader to many recent research results that underscore the effectiveness of GPT-4 and GPT-3. A series of studies, including those by \citet{lee-etal-2020-language}, \citet{press2023measuring}, \citet{pelrine2023reliable}, \citet{yao2023react}, and \citet{jiang2023active}, highlight the effectiveness of GPT models in misinformation detection. \citet{hoes2023using}, in particular, emphasized GPT-3's capability in identifying misinformation in news articles, suggesting that its capabilities extend beyond mere memorization.

To establish the basis of our comparisons in previous research results and set up a reasonable foundation model baseline, we train and test RoBERTa on the \textit{pomt} domain subset of the Multi-FC's training set to demonstrate RoBERTa's competency in misinformation detection. This subset is the most frequent domain in the Multi-FC dataset, containing 15,390 examples. We then compare our foundation model to the top-performing model reported in the original MultiFC results on the similar subset~\cite{augenstein-etal-2019-multifc}.

In this experiment, the input comprises a claim and related evidence snippets retrieved from the Multi-FC dataset. We concatenated these evidence snippets with the claim by separator tokens and tokenizing the resultant string. The model produces an output in the form of a one-hot vector of labels, corresponding to the nine unique classes of the \textit{pomt} domain.

As indicated in Table~\ref{pomt_org}, the RoBERTa-based model demonstrates acceptable performance compared to the previously mentioned Multi-FC model. This is evident from the higher Micro and Macro F1 scores achieved by RoBERTa, confirming its ability to yield satisfactory accuracy levels with our collected dataset and marking it as a suitable baseline for further analysis.

The \textit{pomt} domain consists of 9 labels, some of which contain close to no instances. Due to this label imbalance, the Macro F1 score better represents the model performance. This result indicates that Multi-FC's model is more biased towards classes with many instances, while RoBERTa does equally well on all classes. Note the Multi-FC test set was not made available; thus, the results are reported on the development set and tuned with 10\% of the training set. Given this result, we use RoBERTa in the next experiments and merely focus on the subset that we annotated with persuasive writing strategies.

\begin{table}[!h]
\begin{center}
\begin{tabular}{ |c|c|c| } 
 \hline
  \textbf{Model}  & \textbf{F1$_{Micro}$} & \textbf{F1$_{Macro}$} \\
  \hline
 \small{\cite{augenstein-etal-2019-multifc}}& 0.321    & 0.276 \\ 
 RoBERTa & 0.327    & 0.325 \\ 
 \hline
\end{tabular}
\caption{Multi-FC and RoBERTa-based model results on the \textit{pomt} domain. }
\label{pomt_org}
\end{center}
\end{table}

\subsection{Detection of Persuasive Strategies}
\label{section:persuasive}
We evaluate the prediction of persuasive writing strategies (Task 2) using four RoBERTa-based classifiers for the classification of the labels in various levels of the hierarchy shown in Figure \ref{labels-tree} and explained in Section \ref{section:dataset}. The input to these models is a tokenized text sequence extracted from an article. This text sequence includes a target sentence and is accompanied by nearby sentences in some of our model variations. This format is used for classification models of all levels of the hierarchy. 
The level 1 model is a binary classification model designed to detect whether the input text follows a persuasive writing strategy. The output of level 1 is $\{{Not Persuasive}, Persuasive\}$. In levels 2-4, a piece of text can be annotated with no tags or multiple tags. 
The level 2 classifier includes 12 labels  (e.g., \textit{establishing legitimacy}). The level 3 classifier includes 30 labels  (e.g., \textit{surface credibility marker}). The level 4 classifier model includes nine labels at depth 4 of the label hierarchy (e.g., \textit{medical or scientific jargon}).
 The results of our experiments are shown in Table~\ref{Annotation Results}. We used 193 articles, including 4,249 sentences for training, and the remaining 49 articles, including 1,417 sentences for testing. Also, we ignored labels with a frequency lower than 15 due to label sparsity.
 Our experiments show that extra contextual information generally helps with persuasive strategy labeling, except for level 4. This exception can be justified considering that level 4 strategies are not context-dependent and are usually expressed in one sentence, such as \textit{claiming authority} or \textit{words associated with health}. In contrast, some labels, such as \textit{wrong cause/effect} or sentiments, which are more context-dependent, improve as more context is provided. As mentioned earlier, since our data is highly imbalanced, we consider the Macro-F1 score as the prominent evaluation metric. The detailed evaluation of precision, recall, and F1 per label with the number of training/test examples can be found in Appendix~\ref{appendix:result}.

\begin{table}[h!]
\begin{centering}
\begin{tabular}{ |p{1.5cm}||p{1.5cm}|p{1.5cm}| }
 \hline
 \multicolumn{3}{|c|}{\textbf{Level 1}} \\
 \hline
  \textbf{Context} & \textbf{F1$_{Micro}$} & \textbf{F1$_{Macro}$} \\
 \hline
  None    & 0.668& 0.662  \\ 
 \hline
  Low     & 0.677 & 0.676 \\ 
 \hline
  High    & 0.699 & 0.692  \\ 
 \hline
\end{tabular}

\medskip

\begin{tabular}{ |p{1.5cm}||p{1.5cm}|p{1.5cm}| }
 \hline
 \multicolumn{3}{|c|}{\textbf{Level 2}} \\
 \hline
  \textbf{Context} & \textbf{F1$_{Micro}$} & \textbf{F1$_{Macro}$} \\
 \hline
  None    & 0.468    & 0.342  \\ 
 \hline
  Low     & 0.501    & 0.372 \\
 \hline
  High    & 0.497    & 0.383 \\
 \hline
\end{tabular}

\medskip

\begin{tabular}{ |p{1.5cm}||p{1.5cm}|p{1.5cm}| }
 \hline
 \multicolumn{3}{|c|}{\textbf{Level 3}} \\
 \hline
  \textbf{Context} & \textbf{F1$_{Micro}$} & \textbf{F1$_{Macro}$} \\
 \hline
  None    & 0.430    & 0.270  \\ %
 \hline
  Low     & 0.464    & 0.283 \\ %
 \hline
  High    & 0.463    & 0.288 \\ %
  \hline

 \end{tabular}

\medskip

\begin{tabular}{ |p{1.5cm}||p{1.5cm}|p{1.5cm}| }
 \hline
 \multicolumn{3}{|c|}{\textbf{Level 4}} \\
 \hline
  \textbf{Context} & \textbf{F1$_{Micro}$} & \textbf{F1$_{Macro}$} \\
\hline
  None    & 0.428    & 0.291  \\ %
 \hline
  Low     & 0.434    & 0.265 \\ %
 \hline
  High    & 0.441    & 0.256 \\ %
  \hline
\end{tabular}

\caption{RoBERTa results for four levels of persuasive strategies, with three levels of context.}
\label{Annotation Results}
\end{centering}
\end{table}

\subsection{Misinformation Detection Using Persuasive Strategies}
\label{section:misinfor-with-persuasive}

In this section, we first evaluate misinformation detection on our \textit{Health subset} without using the strategy labels (Task 1) to answer Question 3. 
Then, we evaluate our misinformation detection model with persuasive strategies annotations coming from the prediction of annotation models, i.e., the output of Task 2. In fact, this is Task 3 with a pipeline configuration. In some additional experiments, we use ground-truth strategy labels to obtain the effect of gold annotations when no pipeline errors are propagated from the persuasive strategy labeling step. We specifically evaluate how different input combinations affect the performance, emphasizing whether the inclusion or exclusion of persuasive strategy labels from the articles has any significant impact on the target task.
We fine-tune a RoBERTa-based model on the \textit{Health subset}. The input to this model is a set of tokens that can originate from articles, claims, or persuasive strategy labels, depending on the experiment. In the future, the hierarchical structure of the labels can be exploited as constraints in a more sophisticated approach for modeling this problem and joint training~\cite{rajaby-faghihi-etal-2021-domiknows,RajabFaghihi_Nafar_Zheng_Mirzaee_Zhang_Uszok_Wan_Premsri_Roth_Kordjamshidi_2023}.\\
\noindent\textbf{Noisy and Imbalanced Labels.} 
Our selected subset includes instances from various labeling domains, leading to labels with similar meanings like $Truth!$ and $True$. Such ambiguous labeling can hinder effective training. To address this, we normalized the labels, reducing the initial ten labels to two labels, $True$ and $False$. Detailed information about our label normalization approach is provided in Appendix~\ref{appendix:label-norm}. 
Moreover, we excluded articles from the Multi-FC labeled as $Mixed$. Post-normalization, our subset comprises \textbf{170} instances labeled as $False$ (indicating misinformation) and \textbf{71} instances labeled as $True$ (indicating truthful information). The resulting data elements include a binary label and are used in a binary classification setting. Also, to manage the label imbalance, we integrated a weighted loss into our model.
\\\noindent\textbf{Misinformation Detection without Persuasive Strategy Labels.} 
We first trained and tested RoBERTa using the original article's claim and article to obtain a baseline over our collected health subset. 
Table \ref{Detection Results_1} shows the performance of our models is considerably higher than the majority baseline in terms of the F1-score.

\begin{table}[ht!]
\begin{centering}
\begin{tabular}{ |l|c|c| }
  \hline
 \textbf{Input Source} & \textbf{F1$_{Micro}$} & \textbf{F1$_{Macro}$} \\
  \hline
 Majority Baseline & 0.673 & 0.402  \\ 
  \hline
  \hline
 RoBERTa $_{claim}$& 0.830 & 0.793 \\ 
  \hline
 RoBERTa $_{article}$& 0.810 & 0.765 \\ 
  \hline
RoBERTa $_{claim + article}$& 0.823 & 0.799 \\ 
  \hline
  \hline
   GPT-4 $_{claim}$ & 0.837 & 0.791  \\
  \hline
      GPT-4 $_{article}$ & 0.848 & 0.830  \\
  \hline
    GPT-4 $_{claim + article }$ & \textbf{0.913} & \textbf{0.904}  \\
  \hline
\end{tabular}
\caption{Misinformation Detection results (Task 1). Model$_{source}$ represents the foundation model classifier with $source$ as the input source.}
\label{Detection Results_1}
\end{centering}
\end{table}

 \noindent\textbf{Misinformation Detection with Pipeline of Persuasive Strategy Labels.} In this experiment, we assess the performance of our misinformation detection model when supplemented with persuasive strategy labels. This pipeline configuration begins with applying the persuasive strategy type classifiers (levels 2-4) to the article. Following this, the resulting strategy labels, alongside the articles, are passed to the misinformation detection model. In this configuration, the model outputs are represented as a token-separated string, and the article text was summarized using the Gensim summarizer~\cite{gensim} so that the article text and all persuasive strategies could fit into the RoBERTa input layer with the 512 tokens limit. Alternatively, we could trim the documents and limit the input to the first 512 tokens. However, this setting was suboptimal as we would have lost a large portion of the potentially important information. We did not have such a limitation for the GPT experiments. We evaluated our system with different input source combinations. The results are shown in Table~\ref{Detection Results}.

\begin{table}[h!]
\begin{centering}
\begin{tabular}{ |l|c|c| }
 \hline
 \textbf{Input Source} & \textbf{F1$_{Micro}$} & \textbf{F1$_{Macro}$} \\
  \hline
 GT & 0.898 & 0.872 \\ 
 \hline
  Pred$_{none}$ & 0.687 & 0.622\\ %
 \hline
  Pred$_{low}$ & 0.735 & 0.704 \\ %
 \hline
   Pred$_{high}$ & 0.741 & 0.690 \\ %
 \hline
 Claim+GT &  0.912 & 0.891 \\ 
 \hline
  Claim+Pred$_{none}$  & 0.694 & 0.646 \\ 
 \hline
  Claim+Pred$_{low}$  & 0.728 & 0.683 \\ 
 \hline
  Claim+Pred$_{high}$  & 0.735 & 0.672 \\ 
 \hline
Claim+Article+GT & 0.871 & 0.845 \\ 
 \hline
Claim+Article+Pred$_{none}$ & 0.748 & 0.721 \\ 
 \hline
Claim+Article+Pred$_{low}$ & 0.762 & 0.722 \\ 
 \hline
Claim+Article+Pred$_{high}$ & 0.782 & 0.749 \\ 
  \hline
  \hline
  GPT-4 $_{claim+Pred}$ & 0.783 & 0.681  \\
 \hline
  GPT-4 $_{claim+GT}$ & 0.878 & 0.850  \\
 \hline
  GPT-4 $_{claim+article+Pred}$ & 0.913 & 0.901  \\
 \hline
 GPT-4 $_{claim+article+GT}$ & \textbf{0.939} & \textbf{0.933}  \\
 \hline
 \hline
 GPT-4 $_{in-context \enskip learning}$ & 0.932 & 0.922  \\
 \hline

\end{tabular}
\caption{Misinformation Detection results with different sources (Task 3). Pred$_{context}$ refers to persuasive strategy prediction labels concerning the context level, and GT refers to the persuasive strategy ground truth labels. Articles in the RoBERTa experiments are summarized to satisfy its 512 token limit. GPT-4 $_{in-context \enskip learning}$ result has persuasive strategies information in the prompt.}
\label{Detection Results}
\end{centering}
\end{table}

\begin{figure}[h!]
    \centering
    \fbox{\includegraphics[width=0.44\textwidth]{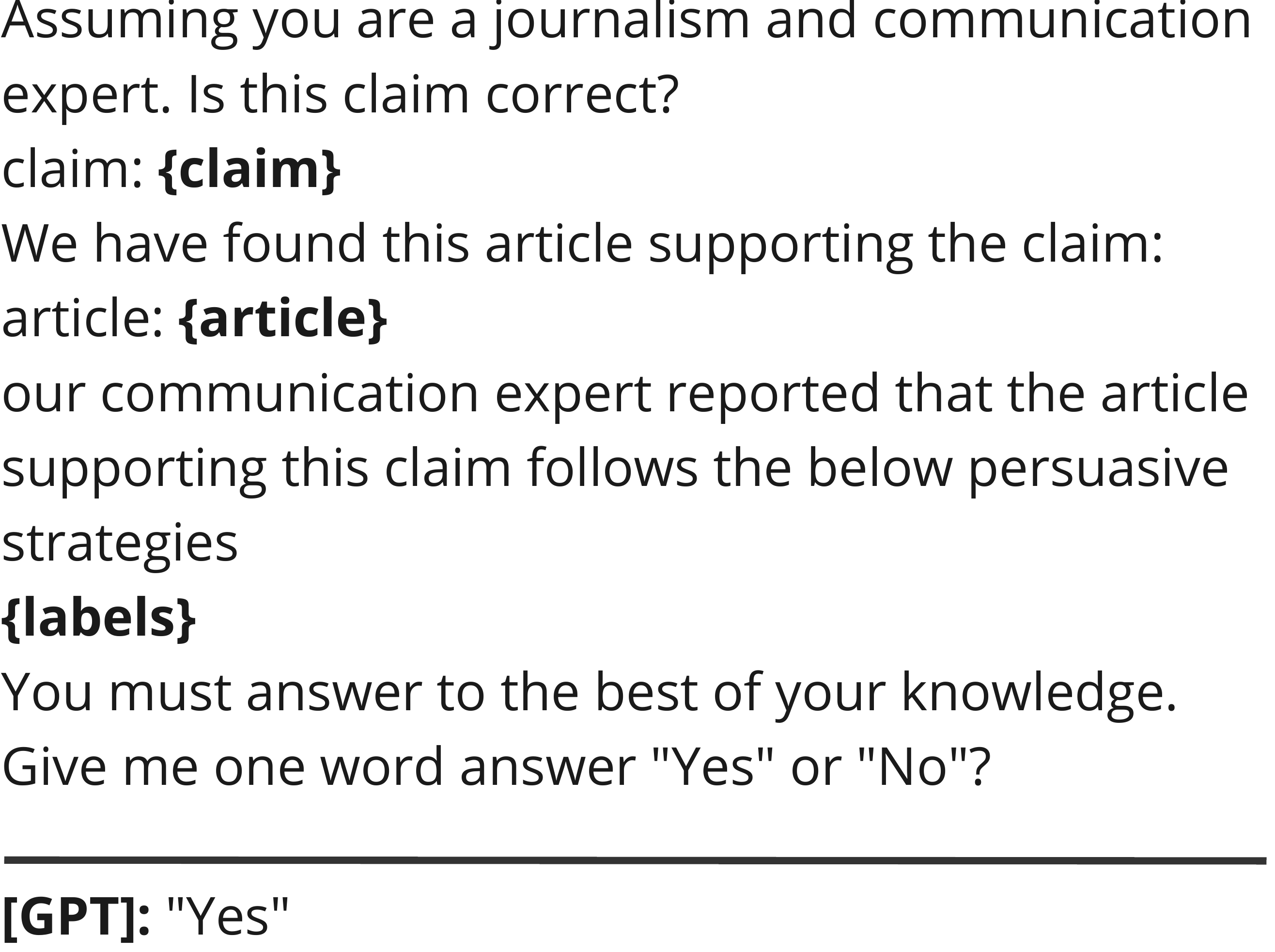}}
    \caption{An example of misinformation detection prompt for GPT model in a zero-shot setting. The prompt structure would vary based on the input sources. More prompt templates can be found in Appendix \ref{appendix:gpt-experiments}.}
    \label{fig:prompt}
\end{figure}

\noindent\textbf{Zero-shot GPT-4.} To explore the ability of generative large language models such as the GPT family on the misinformation detection task, we evaluated GPT-4 in a zero-shot setting using our test set. Similar to the RoBERTa experiment, we evaluate variations of input sources in this experiment by concatenating the information to the zero-shot. We structure our prompt in a way that asks the model to classify the claim based on auxiliary information such as $Article$ or $Labels$. Figure \ref{fig:prompt} shows the structure of our prompt for this experiment. The results of this experiment are shown in Table~\ref{Detection Results_1} and~\ref{Detection Results}. The extension of these results, including the GPT-3 evaluation, can be found in Appendix \ref{appendix:gpt-experiments}.

 \noindent\textbf{Input Variation Observation.} One experimental observation is that incorporating multiple input sources can boost the accuracy of misinformation detection. Our results demonstrate that, in many cases, adding inputs such as $Article$ or $Preds$ to the model enhances its performance compared to using the $Claim$ alone.  While this is the case for GPT, sometimes, the additional information reduces the F1 score for the RoBERTa model. We associate this performance discrepancy with RoBERTa's limited ability to process larger text. Additionally, there might be adverse effects due to the text summarization process, particularly when considering GPT models' capacity to understand longer context and dependencies. This hypothesis becomes even more apparent when we compare the outcomes for $Claim+GT$ with those for $Claim+Article+GT$, where the addition of the $Article$ input led to a marked decrease in RoBERTa's performance.
 
Moreover, the findings depicted in Table~\ref{Detection Results} demonstrate that simply integrating persuasive strategy labels into the prompt of a GPT model can enhance accuracy when operating with cutting-edge large language models. These outcomes indicate the benefits of incorporating persuasive strategies into language models across various sizes and training frameworks.

\begin{figure*}[t!]
\centering
\begin{center}
\begin{tabular}{ |p{420pt}| }
 \hline
\textbf{User:} Here we show example of persuasive strategy detection. \\ 
Examples below show text spans with their corresponding \\ persuasive strategy: \\
span$_1$ $\Rightarrow$ ['persuasive strategy$_1$','persuasive strategy$_2$,..] \\
span$_2$ $\Rightarrow$ ['persuasive strategy$_1$','persuasive strategy$_2$,..]  \\
...\\
Mark the sentences in the text below with persuasive strategy labels: \{article\} \\
\textbf{GPT-4:} [\textit{Then after getting the labels from the GPT-4 model}]  \\
\textbf{User:} Given the labeled persuasive strategies in the above supporting article, Do you think this claim is correct?\\
Claim: The Government is banning organic Farming \\
Give me a Yes or No answer with your chain of thoughts.\\
\textbf{GPT-4:} No, the claim that the government is banning organic farming is not correct based on the labeled persuasive strategies in the supporting article. The article uses a lot of fear and uncertainty to make its point, often exaggerating the potential effects of the bill. It also shows a distrust of government and corporations, suggesting a bias in the information presented. Furthermore, the call to action to read the legislation for oneself implies that the author may not have provided a full or accurate summary of the bill. Therefore, the claim seems to be based more on speculation and fear-mongering than on concrete evidence.\\
 \hline
\end{tabular}
\end{center}
\caption{The prompt structure for the explained misinformation detection using persuasive strategies with an example response.}
\label{example:explaination}
\end{figure*}

\noindent\textbf{Using Ground Truth Strategy Labels.} 
We observed that including ground truth persuasive strategies, or $GT$, markedly bolsters the accuracy of misinformation detection. This is clearly illustrated when comparing the results of $Claim+GT$ and $Claim+Article+GT$ in Table~\ref{Detection Results} with their counterpart experiments without $GT$ in Table~\ref{Detection Results_1}. For instance, GPT-4$_{claim+article+GT}$ outperforms GPT-4$_{claim+article}$ about 2.9\% Macro F1-score. This substantial improvement emphasizes these annotations' significance to the misinformation detection task. On the other hand, the results of the predicted persuasive strategy do not show the same level of performance, both independently and when integrated with the claim or article. As shown in Table \ref{Detection Results}, persuasive strategy labeling quality directly correlates with misinformation detection quality. This implies that augmenting models with high-accuracy persuasive strategy labels can indeed help with misinformation detection, while incorporating noisy strategy labels could, in fact, be detrimental to the model's performance. Further detailed results are demonstrated in Appendix \ref{appendix:mis-examples}.

\noindent\textbf{Two-step Misinformation Detection with predicted Persuasive Strategies in context.}
Here, our aim is to evaluate the real-world applicability of the persuasive strategy labels method by assessing the in-context learning capabilities using the GPT-4 model. To this end, we first select one example from each of our persuasive strategies and the corresponding article span, e.g., "No Matter What" $\Rightarrow$ ['Exaggeration/absolute language,'].
Given this context, we ask the GPT model to identify the persuasive strategies employed within a given article.
After acquiring the labeled strategies, we ask the model to determine the veracity of the claim based on known facts and the marked persuasive writing strategies. As evident from Table \ref{Detection Results}, this step-by-step in-context learning approach enhances the misinformation detection performance compared to zero shot. Even with imperfect labeling of persuasive strategies, we observed a consistent improvement when leveraging them in the prompt.  This result further indicates the effectiveness of the persuasive strategies in both fin-tuning and in-context learning with language models. The prompts we used for this experiment are shown in Appendix \ref{appendix:gpt-experiments}. 
We evaluated the intermediate strategy labeling with GPT, focusing on the second level of the hierarchy that includes labels such as \textit{Narrative with Details} and \textit{Emotional Appeals}. Our results demonstrated an average F1-score of 0.429 in detecting persuasive writing techniques within an article. This result indicates the challenge of accurately detecting the strategies even though it was positively impacting our in-context learning. For further information regarding this experiment, please refer to Appendix \ref{appendix:persuasive-strategies}.

It is worth mentioning that generative models like GPT-4 and GPT-3 offer enhanced explainability. By employing the two-step and CoT prompting in the misinformation detection phase. The detected strategies can be used to explain the reasoning behind the decision that is tailored for the end-user. An example of this is shown in Figure \ref{example:explaination}.

\noindent\textbf{Experiments with a More Recent Dataset.} To further evaluate the effectiveness of the persuasive strategies when used as in-context learning, we employed the RAWFC dataset~\cite{yang-etal-2022-coarse}, which is widely used for in-context learning experiments. The RAWFC dataset is a fact-checking benchmark, providing claims categorized under a three-class scheme: $True$, $False$, or $Half$. This dataset source is real-world news, making it an authentic representation of the claims circulating in today's media landscape. It comprises 200 claims for each training, validation, and testing split. Of these, 67 claims are labeled as $True$, another 67 as $False$, and the remaining 66 are categorized as $Half$ for each split. It also provides at least one article for each claim. For this experiment, first, we chose the two longest articles (if available) of the claim from the dataset. Second, we ask our model to identify persuasive strategies with in-context samples similar to the previous experiment. Subsequently, we ask the model to determine the veracity of the claim provided. The specific prompts utilized for these experiments can be found in Appendix \ref{appendix:gpt-experiments}.
As shown in Table \ref{tab:rawfc}, even using noisy persuasive strategies within the in-context learning setting not only outperforms the standard prompting and Chain-of-thought techniques but also exceeds the accuracy of methods that access online search engine \cite{zhang2023llmbased,yao2023react}. This is a strong result indicating the advantage of using LLM's capability for recognizing communication strategies in language and employing them to inform users about the potential risks of misinformation.

\begin{table}[h]
    \centering
    
    \begin{tabular}{|l|c|c|c|}
        \hline
        Model & P (\%) & R (\%) & F1 (\%) \\
        \hline
        Standard & 48.5 & 48.5 & 48.5 \\
        Standard$_{with articles}$ & 48.5 & 47.5 & 47.9 \\
        CoT  & 42.4 & 46.6 & 44.4 \\
        ReAct$^\dagger$ & 51.2 & 48.5 & 49.8 \\
        HiSS$^\dagger$   & 53.4 & 54.4 & 53.9 \\
        ICL$_{ours}$ & \textbf{56.9} & \textbf{55.5} & \textbf{55.8} \\
        \hline
    \end{tabular}
    \caption{Experimental results of claim verification on the RAWFC. Bold denotes the best performance. $\dagger$ denotes the access to search engines. The results on the RAWFC dataset are brought from the HiSS \cite{zhang2023llmbased}. We used the davinci-003 to ensure a fair comparison.}
    \label{tab:rawfc}
\end{table}

\section{Conclusion}
This study highlights the positive effect of persuasive strategy labeling in misinformation detection, which also helps the need for interpretable results in misinformation detection models to enhance transparency and public trust. We introduce a novel annotation scheme, which categorizes numerous persuasive writing strategies recognized by communication experts and presents a human-annotated dataset. Leveraging large language models, we evaluate various paradigms of fine-tuning and in-context learning to automate strategy detection while discussing associated challenges. We demonstrate a state-of-the-art misinformation detection model that provides explainability without requiring access to external knowledge. In addition, our dataset and the proposed challenge offer valuable resources for future research, aiming to foster critical thinking about online news articles by analyzing their employed writing strategies. By doing so, we strive to contribute towards a more informed populace.

\section{Limitations}

This study has notable limitations. Our primary aim was to analyze the influence of persuasive strategies on competitive baseline models. However, our dataset expansion has some challenges. Annotating persuasive writing strategies demands both expertise and substantial resources, which constrained the size of our dataset. Additionally, the costs associated with GPT's API calls, especially those involving long contexts, are significant and limit the dataset size used for the experiments.
Consequently, we opted for RAWFC, a widely used benchmark with a manageable size. Furthermore, the inherent limitations of automated misinformation detection models should not be overlooked. While they offer valuable insights, and inform the user about the risks, they are not infallible. Their primary role is to supplement, not supplant, human judgment in addressing this critical societal issue. Lastly, the technical demands of large language models might restrict their applicability. Large pre-trained language models require significant computational power, potentially rendering them unsuitable for devices with limited resources.
\section{Acknowledgement}

We are grateful to our anonymous reviewers and Dr. Christos Christodoulopoulos for their invaluable suggestions and discussions. Support for this project was in part provided by the Trifecta Initiative Interdisciplinary Health Research at Michigan State University.\\
\textbf{Credit Statement: }\textbf{DK}: Formal Analysis, Methodology, Software, Experimentation, Writing-Original Draft, Writing-Review and Editing; \textbf{JR}: Formal Analysis, Methodology, Software, Experimentation, Writing-Original Draft; \textbf{HL}: Data Curation, Investigation; \textbf{WP}: Conceptualization, Data Curation, Funding Acquisition, Writing-Original Draft; \textbf{JM}: Conceptualization, Data Curation; \textbf{PK}: Conceptualization, Funding Acquisition, Experimental Design, Supervision, Writing-Review and Editing.

\bibliographystylelanguageresource{lrec-coling2024-natbib}
\bibliographylanguageresource{languageresource}
\bibliographystyle{lrec-coling2024-natbib}
\section{Bibliographical References} 
\bibliography{references}

\begin{thebibliography}{0}
\expandafter\ifx\csname natexlab\endcsname\relax\def\natexlab#1{#1}\fi

\end{thebibliography}


\begin{thebibliography}{70}
\expandafter\ifx\csname natexlab\endcsname\relax\def\natexlab#1{#1}\fi

\bibitem[{Allport and Lepkin(1945)}]{allport1945wartime}
Floyd~H Allport and Milton Lepkin. 1945.
\newblock Wartime rumors of waste and special privilege: Why some people believe them.
\newblock \emph{The Journal of Abnormal and Social Psychology}, 40(1):3.

\bibitem[{Aly et~al.(2021)Aly, Guo, Schlichtkrull, Thorne, Vlachos, Christodoulopoulos, Cocarascu, and Mittal}]{https://doi.org/10.48550/arxiv.2106.05707}
Rami Aly, Zhijiang Guo, Michael Schlichtkrull, James Thorne, Andreas Vlachos, Christos Christodoulopoulos, Oana Cocarascu, and Arpit Mittal. 2021.
\newblock \href {https://doi.org/10.48550/ARXIV.2106.05707} {Feverous: Fact extraction and verification over unstructured and structured information}.

\bibitem[{Atanasova et~al.(2020)Atanasova, Simonsen, Lioma, and Augenstein}]{atanasova2020generating}
Pepa Atanasova, Jakob~Grue Simonsen, Christina Lioma, and Isabelle Augenstein. 2020.
\newblock Generating fact checking explanations.
\newblock \emph{arXiv preprint arXiv:2004.05773}.

\bibitem[{Augenstein et~al.(2019)Augenstein, Lioma, Wang, Chaves~Lima, Hansen, Hansen, and Simonsen}]{augenstein-etal-2019-multifc}
Isabelle Augenstein, Christina Lioma, Dongsheng Wang, Lucas Chaves~Lima, Casper Hansen, Christian Hansen, and Jakob~Grue Simonsen. 2019.
\newblock \href {https://doi.org/10.18653/v1/D19-1475} {{M}ulti{FC}: A real-world multi-domain dataset for evidence-based fact checking of claims}.
\newblock In \emph{Proceedings of the 2019 Conference on Empirical Methods in Natural Language Processing and the 9th International Joint Conference on Natural Language Processing (EMNLP-IJCNLP)}, pages 4685--4697, Hong Kong, China. Association for Computational Linguistics.

\bibitem[{Baly et~al.(2020)Baly, Da~San~Martino, Glass, and Nakov}]{baly-etal-2020-detect}
Ramy Baly, Giovanni Da~San~Martino, James Glass, and Preslav Nakov. 2020.
\newblock \href {https://doi.org/10.18653/v1/2020.emnlp-main.404} {We can detect your bias: Predicting the political ideology of news articles}.
\newblock In \emph{Proceedings of the 2020 Conference on Empirical Methods in Natural Language Processing (EMNLP)}, pages 4982--4991, Online. Association for Computational Linguistics.

\bibitem[{Basol et~al.(2021)Basol, Roozenbeek, Berriche, Uenal, McClanahan, and van~der Linden}]{doi:10.1177/20539517211013868}
Melisa Basol, Jon Roozenbeek, Manon Berriche, Fatih Uenal, William~P. McClanahan, and Sander van~der Linden. 2021.
\newblock \href {https://doi.org/10.1177/20539517211013868} {Towards psychological herd immunity: Cross-cultural evidence for two prebunking interventions against covid-19 misinformation}.
\newblock \emph{Big Data \& Society}, 8(1):20539517211013868.

\bibitem[{Bird et~al.(2009)Bird, Klein, and Loper}]{nltk}
Steven Bird, Ewan Klein, and Edward Loper. 2009.
\newblock \emph{Natural language processing with Python: analyzing text with the natural language toolkit}.
\newblock " O'Reilly Media, Inc.".

\bibitem[{Bommasani et~al.(2022)Bommasani, Hudson, Adeli, Altman, Arora, von Arx, Bernstein, Bohg, Bosselut, Brunskill, Brynjolfsson, Buch, Card, Castellon, Chatterji, Chen, Creel, Davis, Demszky, Donahue, Doumbouya, Durmus, Ermon, Etchemendy, Ethayarajh, Fei-Fei, Finn, Gale, Gillespie, Goel, Goodman, Grossman, Guha, Hashimoto, Henderson, Hewitt, Ho, Hong, Hsu, Huang, Icard, Jain, Jurafsky, Kalluri, Karamcheti, Keeling, Khani, Khattab, Koh, Krass, Krishna, Kuditipudi, Kumar, Ladhak, Lee, Lee, Leskovec, Levent, Li, Li, Ma, Malik, Manning, Mirchandani, Mitchell, Munyikwa, Nair, Narayan, Narayanan, Newman, Nie, Niebles, Nilforoshan, Nyarko, Ogut, Orr, Papadimitriou, Park, Piech, Portelance, Potts, Raghunathan, Reich, Ren, Rong, Roohani, Ruiz, Ryan, Ré, Sadigh, Sagawa, Santhanam, Shih, Srinivasan, Tamkin, Taori, Thomas, Tramèr, Wang, Wang, Wu, Wu, Wu, Xie, Yasunaga, You, Zaharia, Zhang, Zhang, Zhang, Zhang, Zheng, Zhou, and Liang}]{foundation}
Rishi Bommasani, Drew~A. Hudson, Ehsan Adeli, Russ Altman, Simran Arora, Sydney von Arx, Michael~S. Bernstein, Jeannette Bohg, Antoine Bosselut, Emma Brunskill, Erik Brynjolfsson, Shyamal Buch, Dallas Card, Rodrigo Castellon, Niladri Chatterji, Annie Chen, Kathleen Creel, Jared~Quincy Davis, Dora Demszky, Chris Donahue, Moussa Doumbouya, Esin Durmus, Stefano Ermon, John Etchemendy, Kawin Ethayarajh, Li~Fei-Fei, Chelsea Finn, Trevor Gale, Lauren Gillespie, Karan Goel, Noah Goodman, Shelby Grossman, Neel Guha, Tatsunori Hashimoto, Peter Henderson, John Hewitt, Daniel~E. Ho, Jenny Hong, Kyle Hsu, Jing Huang, Thomas Icard, Saahil Jain, Dan Jurafsky, Pratyusha Kalluri, Siddharth Karamcheti, Geoff Keeling, Fereshte Khani, Omar Khattab, Pang~Wei Koh, Mark Krass, Ranjay Krishna, Rohith Kuditipudi, Ananya Kumar, Faisal Ladhak, Mina Lee, Tony Lee, Jure Leskovec, Isabelle Levent, Xiang~Lisa Li, Xuechen Li, Tengyu Ma, Ali Malik, Christopher~D. Manning, Suvir Mirchandani, Eric Mitchell, Zanele Munyikwa, Suraj Nair,
  Avanika Narayan, Deepak Narayanan, Ben Newman, Allen Nie, Juan~Carlos Niebles, Hamed Nilforoshan, Julian Nyarko, Giray Ogut, Laurel Orr, Isabel Papadimitriou, Joon~Sung Park, Chris Piech, Eva Portelance, Christopher Potts, Aditi Raghunathan, Rob Reich, Hongyu Ren, Frieda Rong, Yusuf Roohani, Camilo Ruiz, Jack Ryan, Christopher Ré, Dorsa Sadigh, Shiori Sagawa, Keshav Santhanam, Andy Shih, Krishnan Srinivasan, Alex Tamkin, Rohan Taori, Armin~W. Thomas, Florian Tramèr, Rose~E. Wang, William Wang, Bohan Wu, Jiajun Wu, Yuhuai Wu, Sang~Michael Xie, Michihiro Yasunaga, Jiaxuan You, Matei Zaharia, Michael Zhang, Tianyi Zhang, Xikun Zhang, Yuhui Zhang, Lucia Zheng, Kaitlyn Zhou, and Percy Liang. 2022.
\newblock \href {http://arxiv.org/abs/2108.07258} {On the opportunities and risks of foundation models}.

\bibitem[{Brown et~al.(2020)Brown, Mann, Ryder, Subbiah, Kaplan, Dhariwal, Neelakantan, Shyam, Sastry, Askell, Agarwal, Herbert{-}Voss, Krueger, Henighan, Child, Ramesh, Ziegler, Wu, Winter, Hesse, Chen, Sigler, Litwin, Gray, Chess, Clark, Berner, McCandlish, Radford, Sutskever, and Amodei}]{gpt}
Tom~B. Brown, Benjamin Mann, Nick Ryder, Melanie Subbiah, Jared Kaplan, Prafulla Dhariwal, Arvind Neelakantan, Pranav Shyam, Girish Sastry, Amanda Askell, Sandhini Agarwal, Ariel Herbert{-}Voss, Gretchen Krueger, Tom Henighan, Rewon Child, Aditya Ramesh, Daniel~M. Ziegler, Jeffrey Wu, Clemens Winter, Christopher Hesse, Mark Chen, Eric Sigler, Mateusz Litwin, Scott Gray, Benjamin Chess, Jack Clark, Christopher Berner, Sam McCandlish, Alec Radford, Ilya Sutskever, and Dario Amodei. 2020.
\newblock \href {http://arxiv.org/abs/2005.14165} {Language models are few-shot learners}.
\newblock \emph{CoRR}, abs/2005.14165.

\bibitem[{Burfoot and Baldwin(2009)}]{burfoot2009automatic}
Clint Burfoot and Timothy Baldwin. 2009.
\newblock Automatic satire detection: Are you having a laugh?
\newblock In \emph{Proceedings of the ACL-IJCNLP 2009 conference short papers}, pages 161--164.

\bibitem[{Chen et~al.(2021)Chen, Xiao, and Mao}]{chen2021persuasion}
Sijing Chen, Lu~Xiao, and Jin Mao. 2021.
\newblock Persuasion strategies of misinformation-containing posts in the social media.
\newblock \emph{Information Processing \& Management}, 58(5):102665.

\bibitem[{Cuan-Baltazar et~al.(2020)Cuan-Baltazar, Mu{\~n}oz-Perez, Robledo-Vega, P{\'e}rez-Zepeda, and Soto-Vega}]{cuan2020misinformation}
Jose~Yunam Cuan-Baltazar, Maria~Jos{\'e} Mu{\~n}oz-Perez, Carolina Robledo-Vega, Maria~Fernanda P{\'e}rez-Zepeda, and Elena Soto-Vega. 2020.
\newblock Misinformation of covid-19 on the internet: infodemiology study.
\newblock \emph{JMIR public health and surveillance}, 6(2):e18444.

\bibitem[{Da~San~Martino et~al.(2020)Da~San~Martino, Shaar, Zhang, Yu, Barr{\'o}n-Cede{\~n}o, and Nakov}]{da-san-martino-etal-2020-prta}
Giovanni Da~San~Martino, Shaden Shaar, Yifan Zhang, Seunghak Yu, Alberto Barr{\'o}n-Cede{\~n}o, and Preslav Nakov. 2020.
\newblock \href {https://doi.org/10.18653/v1/2020.acl-demos.32} {{P}rta: A system to support the analysis of propaganda techniques in the news}.
\newblock In \emph{Proceedings of the 58th Annual Meeting of the Association for Computational Linguistics: System Demonstrations}, pages 287--293, Online. Association for Computational Linguistics.

\bibitem[{Da~San~Martino et~al.(2019)Da~San~Martino, Yu, Barr{\'o}n-Cede{\~n}o, Petrov, and Nakov}]{da-san-martino-etal-2019-fine}
Giovanni Da~San~Martino, Seunghak Yu, Alberto Barr{\'o}n-Cede{\~n}o, Rostislav Petrov, and Preslav Nakov. 2019.
\newblock \href {https://doi.org/10.18653/v1/D19-1565} {Fine-grained analysis of propaganda in news article}.
\newblock In \emph{Proceedings of the 2019 Conference on Empirical Methods in Natural Language Processing and the 9th International Joint Conference on Natural Language Processing (EMNLP-IJCNLP)}, pages 5636--5646, Hong Kong, China. Association for Computational Linguistics.

\bibitem[{De~Sarkar et~al.(2018)De~Sarkar, Yang, and Mukherjee}]{de-sarkar-etal-2018-attending}
Sohan De~Sarkar, Fan Yang, and Arjun Mukherjee. 2018.
\newblock \href {https://aclanthology.org/C18-1285} {Attending sentences to detect satirical fake news}.
\newblock In \emph{Proceedings of the 27th International Conference on Computational Linguistics}, pages 3371--3380, Santa Fe, New Mexico, USA. Association for Computational Linguistics.

\bibitem[{Della~Vedova et~al.(2018)Della~Vedova, Tacchini, Moret, Ballarin, DiPierro, and de~Alfaro}]{8468301}
Marco~L. Della~Vedova, Eugenio Tacchini, Stefano Moret, Gabriele Ballarin, Massimo DiPierro, and Luca de~Alfaro. 2018.
\newblock \href {https://doi.org/10.23919/FRUCT.2018.8468301} {Automatic online fake news detection combining content and social signals}.
\newblock In \emph{2018 22nd Conference of Open Innovations Association (FRUCT)}, pages 272--279.

\bibitem[{Dimitrov et~al.(2021)Dimitrov, Bin~Ali, Shaar, Alam, Silvestri, Firooz, Nakov, and Da~San~Martino}]{dimitrov-etal-2021-semeval}
Dimitar Dimitrov, Bishr Bin~Ali, Shaden Shaar, Firoj Alam, Fabrizio Silvestri, Hamed Firooz, Preslav Nakov, and Giovanni Da~San~Martino. 2021.
\newblock \href {https://doi.org/10.18653/v1/2021.semeval-1.7} {{S}em{E}val-2021 task 6: Detection of persuasion techniques in texts and images}.
\newblock In \emph{Proceedings of the 15th International Workshop on Semantic Evaluation (SemEval-2021)}, pages 70--98, Online. Association for Computational Linguistics.

\bibitem[{Eckart~de Castilho et~al.(2016)Eckart~de Castilho, M{\'u}jdricza-Maydt, Yimam, Hartmann, Gurevych, Frank, and Biemann}]{eckart-de-castilho-etal-2016-web}
Richard Eckart~de Castilho, {\'E}va M{\'u}jdricza-Maydt, Seid~Muhie Yimam, Silvana Hartmann, Iryna Gurevych, Anette Frank, and Chris Biemann. 2016.
\newblock \href {https://aclanthology.org/W16-4011} {A web-based tool for the integrated annotation of semantic and syntactic structures}.
\newblock In \emph{Proceedings of the Workshop on Language Technology Resources and Tools for Digital Humanities ({LT}4{DH})}, pages 76--84, Osaka, Japan. The COLING 2016 Organizing Committee.

\bibitem[{Eisend and Tarrahi(2022)}]{eisend2022persuasion}
Martin Eisend and Farid Tarrahi. 2022.
\newblock Persuasion knowledge in the marketplace: A meta-analysis.
\newblock \emph{Journal of Consumer Psychology}, 32(1):3--22.

\bibitem[{Fern{\'a}ndez-Luque and Bau(2015)}]{fernandez2015health}
Luis Fern{\'a}ndez-Luque and Teresa Bau. 2015.
\newblock Health and social media: perfect storm of information.
\newblock \emph{Healthcare informatics research}, 21(2):67--73.

\bibitem[{Friestad and Wright(1994)}]{friestad1994persuasion}
Marian Friestad and Peter Wright. 1994.
\newblock The persuasion knowledge model: How people cope with persuasion attempts.
\newblock \emph{Journal of consumer research}, 21(1):1--31.

\bibitem[{Ghanem et~al.(2021)Ghanem, Ponzetto, Rosso, and Pardo}]{DBLP:journals/corr/abs-2101-09810}
Bilal Ghanem, Simone~Paolo Ponzetto, Paolo Rosso, and Francisco M.~Rangel Pardo. 2021.
\newblock \href {http://arxiv.org/abs/2101.09810} {Fakeflow: Fake news detection by modeling the flow of affective information}.
\newblock \emph{CoRR}, abs/2101.09810.

\bibitem[{Ghanem et~al.(2019)Ghanem, Rosso, and Rangel}]{ghanem2019emotional}
Bilal Ghanem, Paolo Rosso, and Francisco Rangel. 2019.
\newblock \href {http://arxiv.org/abs/1908.09951} {An emotional analysis of false information in social media and news articles}.

\bibitem[{Guess et~al.(2020)Guess, Lerner, Lyons, Montgomery, Nyhan, Reifler, and Sircar}]{guess2020digital}
Andrew~M Guess, Michael Lerner, Benjamin Lyons, Jacob~M Montgomery, Brendan Nyhan, Jason Reifler, and Neelanjan Sircar. 2020.
\newblock A digital media literacy intervention increases discernment between mainstream and false news in the united states and india.
\newblock \emph{Proceedings of the National Academy of Sciences}, 117(27):15536--15545.

\bibitem[{Hameleers(2022)}]{hameleers2022separating}
Michael Hameleers. 2022.
\newblock Separating truth from lies: Comparing the effects of news media literacy interventions and fact-checkers in response to political misinformation in the us and netherlands.
\newblock \emph{Information, Communication \& Society}, 25(1):110--126.

\bibitem[{Hoes et~al.(2023)Hoes, Altay, and Bermeo}]{hoes2023using}
Emma Hoes, Sacha Altay, and Juan Bermeo. 2023.
\newblock Using chatgpt to fight misinformation: Chatgpt nails 72\% of 12,000 verified claims.

\bibitem[{Honnibal et~al.(2020)Honnibal, Montani, Van~Landeghem, and Boyd}]{spacy}
Matthew Honnibal, Ines Montani, Sofie Van~Landeghem, and Adriane Boyd. 2020.
\newblock \href {https://doi.org/10.5281/zenodo.1212303} {{spaCy: Industrial-strength Natural Language Processing in Python}}.

\bibitem[{Islam et~al.(2020)Islam, Sarkar, Khan, Mostofa~Kamal, Hasan, Kabir, Yeasmin, Islam, Amin~Chowdhury, Anwar, Chughtai, and Seale}]{islam2020covid}
Md~Saiful Islam, Tonmoy Sarkar, Sazzad~Hossain Khan, Abu-Hena Mostofa~Kamal, S~M~Murshid Hasan, Alamgir Kabir, Dalia Yeasmin, Mohammad~Ariful Islam, Kamal~Ibne Amin~Chowdhury, Kazi~Selim Anwar, Abrar~Ahmad Chughtai, and Holly Seale. 2020.
\newblock {COVID-19-related} infodemic and its impact on public health: A global social media analysis.
\newblock \emph{Am. J. Trop. Med. Hyg.}, 103(4):1621--1629.

\bibitem[{Jiang et~al.(2023)Jiang, Xu, Gao, Sun, Liu, Dwivedi-Yu, Yang, Callan, and Neubig}]{jiang2023active}
Zhengbao Jiang, Frank~F. Xu, Luyu Gao, Zhiqing Sun, Qian Liu, Jane Dwivedi-Yu, Yiming Yang, Jamie Callan, and Graham Neubig. 2023.
\newblock \href {http://arxiv.org/abs/2305.06983} {Active retrieval augmented generation}.

\bibitem[{Kamali and Kordjamshidi(2023)}]{kamali2023syntax}
Danial Kamali and Parisa Kordjamshidi. 2023.
\newblock Syntax-guided transformers: Elevating compositional generalization and grounding in multimodal environments.
\newblock In \emph{GenBench: The first workshop on generalisation (benchmarking) in NLP}, page 130.

\bibitem[{Khanday et~al.(2021)Khanday, Khan, and Rabani}]{khanday2021identifying}
Akib Mohi Ud~Din Khanday, Qamar~Rayees Khan, and Syed~Tanzeel Rabani. 2021.
\newblock Identifying propaganda from online social networks during covid-19 using machine learning techniques.
\newblock \emph{International Journal of Information Technology}, 13(1):115--122.

\bibitem[{Knapp(1944)}]{knapp1944psychology}
Robert~H Knapp. 1944.
\newblock A psychology of rumor.
\newblock \emph{Public opinion quarterly}, 8(1):22--37.

\bibitem[{Kotonya and Toni(2020)}]{kotonya2020explainable}
Neema Kotonya and Francesca Toni. 2020.
\newblock Explainable automated fact-checking for public health claims.
\newblock \emph{arXiv preprint arXiv:2010.09926}.

\bibitem[{Kouzy et~al.(2020)Kouzy, Abi~Jaoude, Kraitem, El~Alam, Karam, Adib, Zarka, Traboulsi, Akl, and Baddour}]{kouzy2020coronavirus}
Ramez Kouzy, Joseph Abi~Jaoude, Afif Kraitem, Molly~B El~Alam, Basil Karam, Elio Adib, Jabra Zarka, Cindy Traboulsi, Elie~W Akl, and Khalil Baddour. 2020.
\newblock Coronavirus goes viral: quantifying the covid-19 misinformation epidemic on twitter.
\newblock \emph{Cureus}, 12(3).

\bibitem[{Lazer et~al.(2018)Lazer, Baum, Benkler, Berinsky, Greenhill, Menczer, Metzger, Nyhan, Pennycook, Rothschild, Schudson, Sloman, Sunstein, Thorson, Watts, and Zittrain}]{lazer2018science}
David M~J Lazer, Matthew~A Baum, Yochai Benkler, Adam~J Berinsky, Kelly~M Greenhill, Filippo Menczer, Miriam~J Metzger, Brendan Nyhan, Gordon Pennycook, David Rothschild, Michael Schudson, Steven~A Sloman, Cass~R Sunstein, Emily~A Thorson, Duncan~J Watts, and Jonathan~L Zittrain. 2018.
\newblock The science of fake news.
\newblock \emph{Science}, 359(6380):1094--1096.

\bibitem[{Lee et~al.(2020)Lee, Li, Wang, Yih, Ma, and Khabsa}]{lee-etal-2020-language}
Nayeon Lee, Belinda~Z. Li, Sinong Wang, Wen-tau Yih, Hao Ma, and Madian Khabsa. 2020.
\newblock \href {https://doi.org/10.18653/v1/2020.fever-1.5} {Language models as fact checkers?}
\newblock In \emph{Proceedings of the Third Workshop on Fact Extraction and VERification (FEVER)}, pages 36--41, Online. Association for Computational Linguistics.

\bibitem[{Lin et~al.(2018)Lin, Hoover, Portillo-Wightman, Park, Dehghani, and Ji}]{lin2018acquiring}
Ying Lin, Joe Hoover, Gwenyth Portillo-Wightman, Christina Park, Morteza Dehghani, and Heng Ji. 2018.
\newblock Acquiring background knowledge to improve moral value prediction.
\newblock In \emph{2018 ieee/acm international conference on advances in social networks analysis and mining (asonam)}, pages 552--559. IEEE.

\bibitem[{Liu et~al.(2019)Liu, Ott, Goyal, Du, Joshi, Chen, Levy, Lewis, Zettlemoyer, and Stoyanov}]{liu2019roberta}
Yinhan Liu, Myle Ott, Naman Goyal, Jingfei Du, Mandar Joshi, Danqi Chen, Omer Levy, Mike Lewis, Luke Zettlemoyer, and Veselin Stoyanov. 2019.
\newblock \href {http://arxiv.org/abs/1907.11692} {Roberta: A robustly optimized bert pretraining approach}.

\bibitem[{Loomba et~al.(2021)Loomba, de~Figueiredo, Piatek, de~Graaf, and Larson}]{loomba2021measuring}
Sahil Loomba, Alexandre de~Figueiredo, Simon~J Piatek, Kristen de~Graaf, and Heidi~J Larson. 2021.
\newblock Measuring the impact of covid-19 vaccine misinformation on vaccination intent in the uk and usa.
\newblock \emph{Nature human behaviour}, 5(3):337--348.

\bibitem[{Luvembe et~al.(2023)Luvembe, Li, Li, Liu, and Xu}]{LUVEMBE2023103354}
Alex~Munyole Luvembe, Weimin Li, Shaohua Li, Fangfang Liu, and Guiqiong Xu. 2023.
\newblock \href {https://doi.org/https://doi.org/10.1016/j.ipm.2023.103354} {Dual emotion based fake news detection: A deep attention-weight update approach}.
\newblock \emph{Information Processing \& Management}, 60(4):103354.

\bibitem[{Manzoor et~al.(2019)Manzoor, Singla, and Nikita}]{manzoor2019fake}
Syed~Ishfaq Manzoor, Jimmy Singla, and Nikita. 2019.
\newblock Fake news detection using machine learning approaches: A systematic review.
\newblock \emph{2019 3rd International Conference on Trends in Electronics and Informatics (ICOEI)}, pages 230--234.

\bibitem[{Margolin et~al.(2018)Margolin, Hannak, and Weber}]{margolin2018political}
Drew~B Margolin, Aniko Hannak, and Ingmar Weber. 2018.
\newblock Political fact-checking on twitter: When do corrections have an effect?
\newblock \emph{Political Communication}, 35(2):196--219.

\bibitem[{Martino et~al.(2020)Martino, Cresci, Barron-Cedeno, Yu, Di~Pietro, and Nakov}]{https://doi.org/10.48550/arxiv.2007.08024}
Giovanni Da~San Martino, Stefano Cresci, Alberto Barron-Cedeno, Seunghak Yu, Roberto Di~Pietro, and Preslav Nakov. 2020.
\newblock \href {https://doi.org/10.48550/ARXIV.2007.08024} {A survey on computational propaganda detection}.

\bibitem[{Nagy and Kapusta(2021)}]{10.1371/journal.pone.0256940}
Kitti Nagy and Jozef Kapusta. 2021.
\newblock \href {https://doi.org/10.1371/journal.pone.0256940} {Improving fake news classification using dependency grammar}.
\newblock \emph{PLOS ONE}, 16(9):1--22.

\bibitem[{OpenAI(2023)}]{openai2023gpt4}
OpenAI. 2023.
\newblock \href {http://arxiv.org/abs/2303.08774} {Gpt-4 technical report}.

\bibitem[{pandas~development team(2020)}]{pandas}
The pandas~development team. 2020.
\newblock \href {https://doi.org/10.5281/zenodo.3509134} {pandas-dev/pandas: Pandas}.

\bibitem[{Paszke et~al.(2019)Paszke, Gross, Massa, Lerer, Bradbury, Chanan, Killeen, Lin, Gimelshein, Antiga, Desmaison, Kopf, Yang, DeVito, Raison, Tejani, Chilamkurthy, Steiner, Fang, Bai, and Chintala}]{pyTorch}
Adam Paszke, Sam Gross, Francisco Massa, Adam Lerer, James Bradbury, Gregory Chanan, Trevor Killeen, Zeming Lin, Natalia Gimelshein, Luca Antiga, Alban Desmaison, Andreas Kopf, Edward Yang, Zachary DeVito, Martin Raison, Alykhan Tejani, Sasank Chilamkurthy, Benoit Steiner, Lu~Fang, Junjie Bai, and Soumith Chintala. 2019.
\newblock \href {http://papers.neurips.cc/paper/9015-pytorch-an-imperative-style-high-performance-deep-learning-library.pdf} {Pytorch: An imperative style, high-performance deep learning library}.
\newblock In \emph{Advances in Neural Information Processing Systems 32}, pages 8024--8035. Curran Associates, Inc.

\bibitem[{Pelrine et~al.(2023)Pelrine, Reksoprodjo, Gupta, Christoph, and Rabbany}]{pelrine2023reliable}
Kellin Pelrine, Meilina Reksoprodjo, Caleb Gupta, Joel Christoph, and Reihaneh Rabbany. 2023.
\newblock \href {http://arxiv.org/abs/2305.14928} {Towards reliable misinformation mitigation: Generalization, uncertainty, and gpt-4}.

\bibitem[{Popat et~al.(2016)Popat, Mukherjee, Str\"{o}tgen, and Weikum}]{10.1145/2983323.2983661}
Kashyap Popat, Subhabrata Mukherjee, Jannik Str\"{o}tgen, and Gerhard Weikum. 2016.
\newblock \href {https://doi.org/10.1145/2983323.2983661} {Credibility assessment of textual claims on the web}.
\newblock In \emph{Proceedings of the 25th ACM International on Conference on Information and Knowledge Management}, CIKM '16, page 2173–2178, New York, NY, USA. Association for Computing Machinery.

\bibitem[{Press et~al.(2023)Press, Zhang, Min, Schmidt, Smith, and Lewis}]{press2023measuring}
Ofir Press, Muru Zhang, Sewon Min, Ludwig Schmidt, Noah~A. Smith, and Mike Lewis. 2023.
\newblock \href {http://arxiv.org/abs/2210.03350} {Measuring and narrowing the compositionality gap in language models}.

\bibitem[{Radford et~al.(2019)Radford, Wu, Child, Luan, Amodei, and Sutskever}]{Radford2019LanguageMA}
Alec Radford, Jeff Wu, Rewon Child, David Luan, Dario Amodei, and Ilya Sutskever. 2019.
\newblock \href {https://api.semanticscholar.org/CorpusID:160025533} {Language models are unsupervised multitask learners}.

\bibitem[{Rajaby~Faghihi et~al.(2021)Rajaby~Faghihi, Guo, Uszok, Nafar, and Kordjamshidi}]{rajaby-faghihi-etal-2021-domiknows}
Hossein Rajaby~Faghihi, Quan Guo, Andrzej Uszok, Aliakbar Nafar, and Parisa Kordjamshidi. 2021.
\newblock \href {https://doi.org/10.18653/v1/2021.emnlp-demo.27} {{D}omi{K}now{S}: A library for integration of symbolic domain knowledge in deep learning}.
\newblock In \emph{Proceedings of the 2021 Conference on Empirical Methods in Natural Language Processing: System Demonstrations}, pages 231--241, Online and Punta Cana, Dominican Republic. Association for Computational Linguistics.

\bibitem[{Rajaby~Faghihi et~al.(2023)Rajaby~Faghihi, Nafar, Zheng, Mirzaee, Zhang, Uszok, Wan, Premsri, Roth, and Kordjamshidi}]{RajabFaghihi_Nafar_Zheng_Mirzaee_Zhang_Uszok_Wan_Premsri_Roth_Kordjamshidi_2023}
Hossein Rajaby~Faghihi, Aliakbar Nafar, Chen Zheng, Roshanak Mirzaee, Yue Zhang, Andrzej Uszok, Alexander Wan, Tanawan Premsri, Dan Roth, and Parisa Kordjamshidi. 2023.
\newblock \href {https://doi.org/10.1609/aaai.v37i8.26143} {Gluecons: A generic benchmark for learning under constraints}.
\newblock \emph{Proceedings of the AAAI Conference on Artificial Intelligence}, 37(8):9552--9561.

\bibitem[{Rani et~al.(2022)Rani, Das, and Bhardwaj}]{https://doi.org/10.1002/cpe.6479}
Neetu Rani, Prasenjit Das, and Amit~Kumar Bhardwaj. 2022.
\newblock \href {https://doi.org/https://doi.org/10.1002/cpe.6479} {Rumor, misinformation among web: A contemporary review of rumor detection techniques during different web waves}.
\newblock \emph{Concurrency and Computation: Practice and Experience}, 34(1):e6479.

\bibitem[{Rashkin et~al.(2017)Rashkin, Choi, Jang, Volkova, and Choi}]{rashkin-etal-2017-truth}
Hannah Rashkin, Eunsol Choi, Jin~Yea Jang, Svitlana Volkova, and Yejin Choi. 2017.
\newblock \href {https://doi.org/10.18653/v1/D17-1317} {Truth of varying shades: Analyzing language in fake news and political fact-checking}.
\newblock In \emph{Proceedings of the 2017 Conference on Empirical Methods in Natural Language Processing}, pages 2931--2937, Copenhagen, Denmark. Association for Computational Linguistics.

\bibitem[{Roozenbeek et~al.(2020)Roozenbeek, Schneider, Dryhurst, Kerr, Freeman, Recchia, Van Der~Bles, and Van Der~Linden}]{roozenbeek2020susceptibility}
Jon Roozenbeek, Claudia~R Schneider, Sarah Dryhurst, John Kerr, Alexandra~LJ Freeman, Gabriel Recchia, Anne~Marthe Van Der~Bles, and Sander Van Der~Linden. 2020.
\newblock Susceptibility to misinformation about covid-19 around the world.
\newblock \emph{Royal Society open science}, 7(10):201199.

\bibitem[{Shahi and Nandini(2020)}]{DBLP:journals/corr/abs-2006-11343}
Gautam~Kishore Shahi and Durgesh Nandini. 2020.
\newblock \href {http://arxiv.org/abs/2006.11343} {Fakecovid - {A} multilingual cross-domain fact check news dataset for {COVID-19}}.
\newblock \emph{CoRR}, abs/2006.11343.

\bibitem[{Shu et~al.(2017)Shu, Sliva, Wang, Tang, and Liu}]{shu2017fake}
Kai Shu, Amy Sliva, Suhang Wang, Jiliang Tang, and Huan Liu. 2017.
\newblock Fake news detection on social media: A data mining perspective.
\newblock \emph{ACM SIGKDD explorations newsletter}, 19(1):22--36.

\bibitem[{Suarez-Lledo and Alvarez-Galvez(2021)}]{info:doi/10.2196/17187}
Victor Suarez-Lledo and Javier Alvarez-Galvez. 2021.
\newblock \href {https://doi.org/10.2196/17187} {Prevalence of health misinformation on social media: Systematic review}.
\newblock \emph{J Med Internet Res}, 23(1):e17187.

\bibitem[{Swire-Thompson and Lazer(2019)}]{swire2019public}
Briony Swire-Thompson and David Lazer. 2019.
\newblock Public health and online misinformation: challenges and recommendations.
\newblock \emph{Annual review of public health}, 41:433--451.

\bibitem[{Vo and Lee(2019)}]{10.1145/3331184.3331248}
Nguyen Vo and Kyumin Lee. 2019.
\newblock \href {https://doi.org/10.1145/3331184.3331248} {Learning from fact-checkers: Analysis and generation of fact-checking language}.
\newblock In \emph{Proceedings of the 42nd International ACM SIGIR Conference on Research and Development in Information Retrieval}, SIGIR'19, page 335–344, New York, NY, USA. Association for Computing Machinery.

\bibitem[{Vosoughi et~al.(2018)Vosoughi, Roy, and Aral}]{vosoughi2018spread}
Soroush Vosoughi, Deb Roy, and Sinan Aral. 2018.
\newblock The spread of true and false news online.
\newblock \emph{science}, 359(6380):1146--1151.

\bibitem[{Wang(2017)}]{DBLP:journals/corr/Wang17j}
William~Yang Wang. 2017.
\newblock \href {http://arxiv.org/abs/1705.00648} {"liar, liar pants on fire": {A} new benchmark dataset for fake news detection}.
\newblock \emph{CoRR}, abs/1705.00648.

\bibitem[{Wang et~al.(2019)Wang, McKee, Torbica, and Stuckler}]{wang2019systematic}
Yuxi Wang, Martin McKee, Aleksandra Torbica, and David Stuckler. 2019.
\newblock Systematic literature review on the spread of health-related misinformation on social media.
\newblock \emph{Social science \& medicine}, 240:112552.

\bibitem[{Wei~Peng and Meng(2023)}]{doi:10.1080/1369118X.2022.2085615}
Sue~Lim Wei~Peng and Jingbo Meng. 2023.
\newblock \href {https://doi.org/10.1080/1369118X.2022.2085615} {Persuasive strategies in online health misinformation: a systematic review}.
\newblock \emph{Information, Communication \& Society}, 26(11):2131--2148.

\bibitem[{Yang et~al.(2022)Yang, Ma, Chen, Lin, Luo, and Chang}]{yang-etal-2022-coarse}
Zhiwei Yang, Jing Ma, Hechang Chen, Hongzhan Lin, Ziyang Luo, and Yi~Chang. 2022.
\newblock \href {https://aclanthology.org/2022.coling-1.230} {A coarse-to-fine cascaded evidence-distillation neural network for explainable fake news detection}.
\newblock In \emph{Proceedings of the 29th International Conference on Computational Linguistics}, pages 2608--2621, Gyeongju, Republic of Korea. International Committee on Computational Linguistics.

\bibitem[{Yao et~al.(2023)Yao, Zhao, Yu, Du, Shafran, Narasimhan, and Cao}]{yao2023react}
Shunyu Yao, Jeffrey Zhao, Dian Yu, Nan Du, Izhak Shafran, Karthik~R Narasimhan, and Yuan Cao. 2023.
\newblock \href {https://openreview.net/forum?id=WE_vluYUL-X} {React: Synergizing reasoning and acting in language models}.
\newblock In \emph{The Eleventh International Conference on Learning Representations}.

\bibitem[{Yuliani et~al.(2019)Yuliani, Sahib, and MFBAF}]{yuliani2019hoax}
Sy~Yuliani, Shahrin Sahib, and ZR~MFBAF. 2019.
\newblock Hoax news classification using machine learning algorithms.
\newblock \emph{International Journal of Engineering and Advanced Technology}, 9(2):3938--3944.

\bibitem[{Zhang and Gao(2023)}]{zhang2023llmbased}
Xuan Zhang and Wei Gao. 2023.
\newblock \href {http://arxiv.org/abs/2310.00305} {Towards llm-based fact verification on news claims with a hierarchical step-by-step prompting method}.

\bibitem[{Řehůřek and Sojka(2010)}]{gensim}
Radim Řehůřek and Petr Sojka. 2010.
\newblock \href {http://www.fi.muni.cz/usr/sojka/presentations/lrec2010-poster-rehurek-sojka.pdf} {Software framework for topic modelling with large corpora}.
\newblock In \emph{Proceedings of LREC 2010 workshop New Challenges for NLP Frameworks}, pages 46--50, Valletta, Malta. University of Malta.

\end{thebibliography}

\appendix

\section{GPT Experiments}
\label{appendix:gpt-experiments}
We evaluate the GPT-4\footnote{gpt-4 model version as of 9/20/2023} and GPT-3 \footnote{text-davinci-003} models in zero-shot and in-context learning setting on the test set of our dataset. We used the prompts shown in Table \ref{tab:prompts} to evaluate models for the misinformation detection task. The labels are the names of the persuasive strategies concatenated with ". " (dot space).
Table \ref{Detection Results GPT} analysis reveals GPT-4's superior performance over GPT-3 in misinformation detection, which could be due to its sophisticated architecture and updated training data. Remarkably, GPT-4 shows a significant improvement when leveraging $GT$ ground truth data, which indicates capability in misinformation detection requiring nuanced context discernment. Its enhanced ability to handle $Claim + Article$ input combinations is pivotal in detecting complex misinformation narratives. Importantly, the best performance for both models is achieved when combining $GT,$ $Claim,$ and $Article$, highlighting the crucial role of reliable ground truth in improving model accuracy.
In conclusion, the observed results indicate a significant advancement in the capacity of language models, evidenced by the progression from GPT-3 to GPT-4. Moreover, the findings reinforce the assertion that the incorporation of persuasive strategies can contribute significantly to the efficiency of misinformation detection.

\begin{table*}[h!]
    \centering
    \begin{tabular}{|c|l|}
    \hline
         \textbf{Inputs} & \textbf{Prompt}  \\
         \hline
         \hline
         \multirow{4}{*}{Zero-shot$_{Claim}$}&  Assuming you are a journalism and communication expert. Is this claim correct? \\
         & claim: \{claim\}\\
         & You must answer to the best of your knowledge. Give me one word answer "Yes"\\
         & or "No"? \\
    
    \hline
         \multirow{4}{*}{Zero-shot$_{Labels}$}&  Assuming you are a journalism and communication expert.  Is a claim correct if \\ 
         & our communication expert reported that an article supporting that claim follows \\ & the below persuasive strategies. \\
         & \{labels\}\\
         & You must answer to the best of your knowledge. Give me one word answer "Yes" \\
         & or "No"? \\
     \hline
         \multirow{4}{*}{Zero-shot$_{Article}$}&  Assuming you are a journalism                              and communication expert. Is this article correct? \\
         & article: \{article\}\\
         & You must answer to the best of your knowledge. Give me one word answer "Yes"\\
         & or "No"? \\
     \hline
         \multirow{7}{*}{Zero-shot$_{Claim}$}&  Assuming you are a journalism                              and communication expert. Is this claim correct? \\
         & claim: \{claim\}\\
         & our communication expert reported that the article supporting this claim follows \\ 
         & the below persuasive strategies \\
        $_{+ Labels}$ & \{labels\} \\
         & You must answer to the best of your knowledge. Give me one word answer "Yes"\\
         & or "No"? \\
     \hline
         \multirow{9}{*}{Zero-shot$_{Claim}$}&  Assuming you are a journalism                              and communication expert. Is this claim correct? \\
         & claim: \{claim\}\\
         & We have found this article supporting the claim: \\
         & article: \{article\} \\
         & our communication expert reported that the article supporting this claim follows \\ 
        $+ _{Article + Labels}$ & the below persuasive strategies \\
        & \{labels\} \\
         & You must answer to the best of your knowledge. Give me one word answer "Yes"\\
         & or "No"? \\
     \hline

          \multirow{13}{*}{In-context Learning}
        & Here we show examples of persuasive strategy detection. \\ 
        &Examples below show text spans with their corresponding persuasive strategy: \\
        & span$_1$ $\Rightarrow$ ['persuasive strategy$_1$','persuasive strategy$_2$,..] \\
        & span$_2$ $\Rightarrow$ ['persuasive strategy$_1$','persuasive strategy$_2$,..]  \\
        GPT& ...\\
        & Mark the sentences in the text below with persuasive strategy labels: \{article\} \\
        & Don't mark a sentence with one strategy more than once.\\
        & [\textit{Then after getting the labels from the GPT model}]  \\
        
        &Given the labeled persuasive strategies in the above-supporting article, is the \\ &following claim correct? Claim: \{claim\}.\\
        & Give me a Yes or No answer with your chain of thoughts. \\
        
     \hline

    \end{tabular}
    \caption{GPT model prompts. $Labels$ refers to either ground-truth persuasive strategy labels ($GT$) or prediction of our Task 2 models ($Pred_{context}$).}
    \label{tab:prompts}
\end{table*}

\begin{table*}[h!]
\begin{centering}
\begin{tabular}{ |l|c|c|c|c| }
  \hline
  \multirow{2}{*}{\textbf{Input Source}}
   & \multicolumn{2}{|c|}{GPT3} & \multicolumn{2}{|c|}{GPT4} \\
  
  & \textbf{F1$_{Micro}$} & \textbf{F1$_{Macro}$} &\textbf{F1$_{Micro}$} & \textbf{F1$_{Macro}$} \\
  \hline
  \hline

\hline
 GT & 0.653 & 0.612 &  0.696 & 0.510 \\ %
 \hline
 Pred$_{high}$ & 0.592 & 0.372 &  0.653 & 0.395 \\ %
 \hline
 Claim &  0.837 & 0.779 & 0.837 & 0.791 \\ %
 \hline
 Article &  0.816 & 0.805 & 0.878 & 0.865 \\ %
 \hline
 Claim + Article &  0.857 & \textbf{0.851} & 0.898 & 0.891 \\ %
 \hline
 Claim + Pred$_{high}$ &  0.755 & 0.686 & 0.776 & 0.688 \\ %
    \hline
 Claim + GT &  0.837 & 0.791 & 0.898 & 0.878 \\ %
    \hline
Claim + Article + Pred$_{high}$ & 0.653 &  0.648 & 0.918 &  0.910\\ %
    \hline
 Claim + Article + GT & \textbf{0.857} & 0.779 & \textbf{0.939} &  \textbf{0.933}\\ %
 \hline

\end{tabular}
\caption{Misinformation Detection results (Task 3) of GPT models with $source$ and prompts shown in Table \ref{tab:prompts}. The results are an average of three runs with 0 temperature.}
\label{Detection Results GPT}
\end{centering}
\end{table*}

\section{Persuasive Strategies Detection using GPT}
\label{appendix:persuasive-strategies}
In this study, we conduct a preliminary assessment of the capability of GPT-3.5 to identify persuasive writing strategies. Using the ChatGPT website interface for our experiment, we first provided the model with the descriptions of the persuasive writing strategies provided in the subsections of "Persuasive strategy themes" in \citet{doi:10.1080/1369118X.2022.2085615}. Subsequently, we tasked GPT-3.5 with detecting the persuasive strategies used in a given text. An example of the interactions is shown in Figure \ref{persuasive-strategy-detection-example:1}. In this experiment, we evaluated the model's performance based solely on its ability to identify the existence of a persuasive writing strategy within the article without delving into specific sentences or segments. We repeated this experiment twice and reported the average accuracy of the GPT-3.5 model on this task in Table \ref{tab:persuasive-gpt}. According to the findings in Table \ref{tab:persuasive-gpt}, GPT-3.5 exhibits better capability in detecting certain strategies, such as \textit{Distinctive Linguistic Features} and \textit{Narrative with Details}, while it faces challenges with more nuanced tasks like identifying \textit{Rhetorical Tricks}, which are inherently difficult to pinpoint. It is important to note that the results summarized in Table \ref{tab:persuasive-gpt} encompass the entirety of the articles included in our dataset. These results further signify the difficulty of persuasive writing strategy detection in automated systems despite being very helpful for analyzing misinformation.

\begin{figure*}[h!]
\centering
\begin{center}
\begin{tabular}{ |p{400pt}| }
 \hline
\textbf{User}: Please learn from the persuasive strategies in online health misinformation below. 
Persuasive strategies are used to create persuasive information or misinformation. [Persuasive strategies Descriptions].\\

\textbf{GPT}: [GPT Response]\\

\textbf{User}: Using the 12 strategies you learned above, identify the persuasive strategies in the following text: [A popular Chinese restaurant in Pretoria has been granted permission to sell Dog Meat to its customers. The restaurant successfully argued that the banning of consuming dog meat, which they pointed out, is in violation of their religious and cultural rights. The restaurant has been granted provisional permission to slaughter, sell and consume dog meat, while awaiting the Supreme Court's final decision of their case. The restaurant is allowed to slaughter up to 35 dogs per week, under the conditions that the dogs are killed properly and humanely. One animal rights activist called the decision "shocking" and vowed to fight the restaurant in court to block the decision. "We can not believe that such actions would be allowed on South African soil. Dogs are precious and should not be used as food. We will not allow the courts to follow through with giving these savages permission to consume man's best friend," said Jon Smit an animal rights activist.]

\textbf{GPT}: From the provided text, the following persuasive strategies are present:
Highlighting Uncertainty and Risk: The text highlights potential ethical and health concerns by discussing the consumption of dog meat and referencing the need to kill dogs "properly and humanely."
Exploiting Science’s Limitations: The text uses the term "properly and humanely" to imply that the practice might not be humane, questioning the treatment of animals.
Inappropriate Use of Scientific Evidence: The claim about the restaurant being allowed to slaughter up to 35 dogs per week is presented without clear scientific evidence or context.
Rhetorical Tricks: The text employs emotive language such as "shocking," "savages," and "man's best friend" to evoke strong emotional reactions and sway opinion.
Emotional Appeals: Emotional language, such as describing dogs as "precious" and "man's best friend," aims to evoke strong emotional responses from the readers.
Establishing Legitimacy: The text cites an animal rights activist named Jon Smit to strengthen the argument against the practice and lend credibility to the opposing viewpoint.
It's important to note that while these strategies are used in the text, the authenticity and accuracy of the information presented are uncertain. Always verify information from reliable sources before believing or sharing claims.
\\
\hline
\end{tabular}
\end{center}
\caption{An example of persuasive strategy detection using GPT-3.5.}
\label{persuasive-strategy-detection-example:1}
\end{figure*}

\begin{table*}[t]
    \begin{tabular}{|l|l|}
    \hline
    \textbf{Persuasive Writing Strategy}  & \textbf{F-1 Score} \\
    \hline
    \hline
    Narrative with Details & 0.534 \\
    \hline
    Using Anecdotes and Personal Experience as Evidence & 0.285 \\
    \hline
    Distrusting Government or Corporations & 0.407 \\
    \hline
    Politicizing Health Issues & 0.341 \\
    \hline
    Highlighting Uncertainty and Risk & 0.365 \\
    \hline
    Exploiting Science’s Limitations & 0.067 \\
    \hline
    Inappropriate Use of Scientific Evidence & 0.477 \\
    \hline
    Rhetorical Tricks & 0.221 \\
    \hline
    Biased Reasoning to Make a Conclusion & 0.457 \\
    \hline
    Emotional Appeals & 0.528 \\
    \hline
    Distinctive Linguistic Features & 0.599 \\
    \hline
    Establishing Legitimacy & 0.869 \\
    \hline
    \textbf{Average} & \textbf{0.429} \\
    \hline
    \end{tabular}%
  \centering
  \caption{GPT-3.5 performance on level-2 persuasive writing strategy detection. The results are at an article level instead of the sentence level compared to Task 2, which is sentence-based. The results are the average of two runs.}
  \label{tab:persuasive-gpt}%
\end{table*}%

\section{Models Hyperparameters}
Models hyperparameters of each experiment are shown in Tables \ref{tab:hyper-persuasive}, \ref{tab:hyper-baseline}, and \ref{tab:hyper-misinformation}.
\label{appendix:hyper}
\begin{table}[h!]
\begin{center}
\begin{tabular}{ |c|c| } 
 \hline
\textbf{Parameter} & \textbf{Value} \\ 
  \hline
  \hline
Batch Size & 20 \\
 \hline
Learning Rate & 5e-5 \\ 
 \hline
RoBERTa Max length & 512  \\ 
  \hline
 Epoch  \#: Level 1 & 40 \\
  Epoch  \#: Level 2-4 & 60 \\
  \hline
 Linear Layer Size: Level 1 & 256 \\
 Linear Layer Size: Level 2 & 1024 \\
 Linear Layer Size: Level 3 & 1024 \\
 Linear Layer Size: Level 4 & 1024 \\
  \hline
 Train/Test Split & 80\% / 20\% \\
 \hline
 Dropout & 50\% \\
 \hline
\end{tabular}
\end{center}
\caption{Persuasive Strategy Models Hyperparameters}
\label{tab:hyper-persuasive}
\label{params}
\end{table}

\begin{table}[h!]
\begin{center}
\begin{tabular}{ |c|c| } 
 \hline
\textbf{Parameter} & \textbf{Value} \\ 
  \hline
  \hline
Batch Size & 10 \\
 \hline
Learning Rate & 5e-5 \\ 
 \hline
RoBERTa Max length & 512  \\ 
  \hline
  Epoch  \# & 12 \\
  \hline
 Linear Layer Size & None \\
  \hline
 Train/Dev Split & 90\% / 10\% \\
 \hline
 Dropout & 50\% \\
 \hline
\end{tabular}
\end{center}
\caption{RoBERTa model's hyperparameters for \textit{pomt} baseline comparison}
\label{tab:hyper-baseline}
\end{table}

\begin{table}[h!]
\begin{center}
\begin{tabular}{ |c|c| } 
 \hline
\textbf{Parameter} & \textbf{Value} \\ 
  \hline
  \hline
Batch Size & 10 \\
 \hline
Learning Rate & 5e-5 \\ 
 \hline
RoBERTa Max length & 512  \\ 
  \hline
  Epoch  \# & 12 \\
  \hline
 Linear Layer Size & None \\
  \hline
 Train/Test Split & 80\%  / 20\% \\
 \hline
 Dropout & 50\% \\
 \hline
  RoBERTa frozen layers \#  & 181 \\
 \hline
\end{tabular}
\end{center}
\caption{Misinformation Detection Model Hyperparameters}
\label{tab:hyper-misinformation}
\end{table}

\section{Misinformation Detection Examples}
\label{appendix:mis-examples}
Figure \ref{example:1} is an example of how the correct persuasive strategy prediction helps with detection. 
In Figure \ref{example:2}, we can see that our persuasive strategy labeler has detected more strategies than the ground truth, which has affected the model's decision to wrongfully mislabel as misinformation. 
Figure \ref{example:3} portrays an example in which wrong strategy prediction did not help the model. However, we can see that if we use ground truth strategy labels, the model can correct its prediction and correctly label the claim.

\begin{figure*}[h!]
\centering
\begin{center}
\begin{tabular}{ |p{400pt}| }
 \hline
\textbf{Claim}: HIV infected needle at McDonald's in the ball pit\newline
\textbf{Article}: Email: (important) McDonalds Playground
Some of you might not be parents, but you may have nieces, nephews, grandchildren or friends with children this will pertain to you too.
As I read the following, my heart sank.
I urge each and every one of you to pass this on to as many people as you can.
I cannot stress how important this is!
This is very disturbing news.
In addition to the following true story, I will also add that my own sons were playing in the ball pit at Discovery Zone one day.
One son lost his watch, and was very upset...\newline
\textbf{GT}:  emotional appeals, call to action, establishing legitimacy, fear, using anecdotes and personal experience as evidence, narrative with details verified to be false, narrative with details, narrative with details verified to be false, narrative with details, highlighting uncertainty and risk, establishing legitimacy, legitimate persuasive techniques: rhetorical question, call to action, establishing legitimacy\newline
\textbf{Pred}: establishing legitimacy, call to action, emotional appeals, shifting hypothesis, using anecdotes and personal experience as evidence, narrative with details, narrative with details details not verified, using anecdotes and personal experience as evidence, using anecdotes and personal experience as evidence, narrative with details, using anecdotes and personal experience as evidence, narrative with details, narrative with details verified to be false, narrative with details, using anecdotes and personal experience as evidence, narrative with details verified to be false, narrative with details, establishing legitimacy, highlighting uncertainty and risk, narrative with details, narrative with details, narrative with details, establishing legitimacy, citing seemingly credible source, legitimate persuasive techniques: rhetorical question, establishing legitimacy, call to action \newline
\textbf{Claim}: No Misinformation \newline
\textbf{Claim + Ground Truth}: Misinformation \newline
\textbf{Claim + Pred}: Misinformation \newline
\textbf{Actual Label}: Misinformation
\\
 \hline
\end{tabular}
\end{center}
\caption{Example of when Persuasive Strategies helped with misinformation Detection }
\label{example:1}
\end{figure*}

\begin{figure*}[h!]
\centering
\begin{center}
\begin{tabular}{ |p{400pt}| } 
 \hline
\textbf{Claim}: In July 2017, it was announced that Finnish researchers would begin human trials of a Type 1 diabetes vaccine in 2018\newline
\textbf{Article}: Finnish diabetes vaccine trials to start in 2018
Finnish researchers have developed an experimental vaccine against type 1 diabetes, and tested it successfully on mice.
An enterovirus, which likely causes type 1 diabetes.
Image: Lääketieteen ja biotieteiden tiedekunta / Tampereen yliopisto
A vaccine for type 1 diabetes developed by Finnish researchers will be tested on mainly Finnish human subjects in late 2018, researchers announced on Tuesday...\newline
\textbf{GT}:  establishing legitimacy, citing source to establish legitimacy source verified to be credible, establishing legitimacy, surface credibility markers - simply claiming authority or credibility, establishing legitimacy, citing source to establish legitimacy source verified to be credible\newline
\textbf{Pred}: surface credibility markers, exaggeration, establishing legitimacy, surface credibility markers, surface credibility markers - simply claiming authority or credibility, establishing legitimacy, less robust evidence or outdated evidence verified, establishing legitimacy, surface credibility markers - simply claiming authority or credibility, establishing legitimacy, ellipses, exaggerated/excessive usage of punctuation marks, citing source to establish legitimacy source not verified, inappropriate use of scientific evidence, lack of citation for evidence, highlighting uncertainty and risk, biased reasoning to make a conclusion, inappropriate analogy or false connection, narrative with details, narrative with details details not verified, establishing legitimacy, inappropriate use of scientific evidence, distinctive linguistic features, anxiety\newline
\textbf{Predictions:}\newline
\textbf{Claim}: Misinformation \newline
\textbf{Claim + Ground Truth}: No Misinformation  \newline
\textbf{Claim + Pred}: Misinformation  \newline
\textbf{Actual Label}: No Misinformation  \\
  \hline

\end{tabular}
\end{center}
\caption{Example of when Persuasive Strategies Prediction did NOT help with the misinformation Detection}
\label{example:2}
\end{figure*}

\begin{figure*}[h!]
\centering
\begin{center}
\begin{tabular}{ |p{400pt}| } 
 \hline

\textbf{Claim}: Doctors are now warning parents to never use baby wipes because they contain a "chemical" called methylisothiazolinone.\newline
\textbf{Article}: Doctor's Warning: Never Clean Your Child With Baby Wipes No Matter What
Many parents wouldn't think twice about what's in their child's baby wipes.
More and more research is surfacing that baby wipes aren't the best thing for our children's skin.
In fact, some doctors are recommending that parents cease using baby wipes on their children entirely because of the risk they are taking each time they do so.
According to NBC News, a study has surfaced that confirms baby wipes are not safe to use on your children due to one key ingredient...\newline
\textbf{GT}:  title of article as clickbait, distinctive linguistic features, exaggeration/absolute language, rhetorical tricks, ellipses, exaggerated/excessive usage of punctuation marks, distinctive linguistic features, establishing legitimacy, legitimate persuasive techniques: rhetorical question, surface credibility markers - simply claiming authority or credibility, establishing legitimacy, fear, emotional appeals, surface credibility markers - simply claiming authority or credibility, establishing legitimacy, rhetorical tricks, exaggeration/absolute language, establishing legitimacy, citing source to establish legitimacy source not verified, highlighting uncertainty and risk, citing source to establish legitimacy source not verified, establishing legitimacy, inappropriate use of scientific and other evidence - out of context verified, inappropriate use of scientific and other evidence, surface credibility markers - words associated with nature or healthiness, establishing legitimacy, establishing legitimacy, citing source to establish legitimacy source not verified, rhetorical tricks, exaggeration/absolute language, highlighting uncertainty and risk, hope, emotional appeals\newline
\textbf{Pred}: distinctive linguistic features, bolded words or underline, establishing legitimacy, citing seemingly credible source, establishing legitimacy, surface credibility markers, establishing legitimacy, surface credibility markers - simply claiming authority or credibility, establishing legitimacy, ellipses, exaggerated/excessive usage of punctuation marks, surface credibility markers, establishing legitimacy, ellipses, exaggerated/excessive usage of punctuation marks, citing source to establish legitimacy source not verified, narrative with details, narrative with details details not verified, narrative with details, narrative with details details verified to be true, narrative with details, narrative with details, narrative with details, establishing legitimacy, narrative with details details not verified, narrative with details, ellipses, exaggerated/excessive usage of punctuation marks, narrative with details, narrative with details details not verified, establishing legitimacy, ellipses, exaggerated/excessive usage of punctuation marks, citing source to establish legitimacy source not verified\newline
\textbf{Predictions:}\newline
\textbf{Claim}: Misinformation \newline
\textbf{Claim + Ground Truth}: Misinformation  \newline
\textbf{Claim + Pred}: No Misinformation  \newline
\textbf{Actual Label}: Misinformation  \\
 \hline
\end{tabular}
\end{center}
\caption{Example of Persuasive Strategies Prediction when it worsened the Detection}
\label{example:3}
\end{figure*}

\section{Dataset Statistics}
\label{appendix:data}
Tables \ref{parent-child} and \ref{child-sub} present the distribution statistics of sentences marked with labels shown in Figure \ref{labels-tree} considering their parent-children relationship. It should be mentioned that a parent is not necessarily tagged with any of its children's tags, while a child tag must always be accompanied by its parent tag.

\begin{table*}[h!]
\begin{centering}
\begin{tabular}{ |l|l|l| } 
\hline
\multicolumn{3}{|c|}{\textbf{Full Annotation Table}} \\
\hline
\hline
\textbf{Parent} & \textbf{Child} & \textbf{\# of Instances} \\
\hline
\multirow{1}{10em}{Narrative w/ Details} & Total & 555 \\\cline{2-3} 
& Verified to be False & 253 \\ 
& Verified to be True & 182 \\ 
& Not Verified & 91 \\
\hline
\multirow{3}{10em}{Anecdotes and Personal Experience as Evidence} & Total & 239 \\
 &  & \\
 &  & \\
\hline
\multirow{1}{10em}{Distrusting Govt. or Corporations} & Total & 104 \\\cline{2-3} 
& Financial Motive & 64 \\ 
\hline
\multirow{1}{10em}{Politicizing Health Issues} & Total & 54 \\\cline{2-3}
& Freedom of Choice and Agency & 12 \\
& Ingroup vs. Outgroup & 16\\ 
& Political Figures/ Argument & 13 \\ 
& Religion or Ideology & 10 \\ 
\hline
Highlighting Uncertainty or Risk & Total & 177\\ 
\hline
Exploiting Science's Limitations & Total & 7\\ 
\hline
\multirow{1}{12em}{Inappropriate use of Scientific or other Evidence} & Total & 294\\\cline{2-3}
& Out of Context/ Verified & 117\\ 
& Less robust or outdated Evidence/ Verify & 51\\ 
\hline
\multirow{1}{10em}{Rhetorical Tricks} & Total & 60 \\\cline{2-3}
& Exaggeration/ Absolute Language & 50\\ 
& Selective Omission & 0\\ 
\hline
\multirow{2}{10em}{Biased Reasoning to make a Conclusion} & Total & 382\\\cline{2-3}
& Inappropriate Analogy or False connection & 43 \\ 
& Wrong Cause/Effect & 18\\ 
& Lack of Evidence or Incomplete Evidence & 247\\ 
& Evidence does not support Conclusion & 34 \\ 
& Shifting Hypothesis & 32\\ 
\hline
\multirow{1}{10em}{Emotional Appeals} & Total & 208 \\\cline{2-3}
& Fear & 98\\ 
& Anger & 40\\ 
& Hope & 19\\ 
& Anxiety & 36\\ 
\hline
\multirow{2}{10em}{Distinctive Linguistic Features} & Total & 516 \\\cline{2-3}
& Uppercase Words & 233\\ 
& Linguistic Intensifier & 1\\ 
& Clickbait Title & 47\\ 
& Bolded, underlined, or italicized content & 143\\ 
& Excessive usage of Punctuation Marks & 131\\ 
\hline
\multirow{1}{10em}{Establishing Legitimacy} & Total & 1045\\\cline{2-3}
& Citing Source To Establish Legitimacy & 399\\ 
& Legitimate Persuasive Techniques & 191\\ 
& Surface Credibility Markers & 346\\ 
& Call to Action & 132 \\ 
\hline

\end{tabular}
\caption{Exhaustive Table of all Parent and Child Annotations, along with the number of examples annotated for each.}
\label{parent-child}
\end{centering}
\end{table*}


\begin{table*}[h!]
\begin{centering}
\begin{tabular}{ |l|l|l|l| } 
\hline
\multicolumn{4}{|c|}{\textbf{Sub Tag Table}} \\
\hline
\hline
\textbf{Parent} & \textbf{Child} & \textbf{Sub} & \textbf{\# of Instances}\\
\hline
\multirow{13}{6em}{Establishing} & \multirow{2}{10em}{Citing Source To Establish Legitimacy} & Total & 399 \\\cline{3-4}
& & Verified to be Credible & 86\\ 
& & Verified to not be credible & 29\\ 
& & Not Verified & 274 \\ 
& & Verified to be Made Up & 8 \\\cline{2-4}
& \multirow{2}{10em}{Legitimate Persuasive Techniques} & Total & 191 \\\cline{3-4}
& & Rhetorical Question & 181 \\ 
Legitimacy& & Metaphor & 0\\ 
& & Humor & 8\\\cline{2-4}
& \multirow{2}{10em}{Surface Credibility Markers} & Total & 346 \\\cline{3-4}
& & Medical/Scientific Jargon & 76 \\ 
& & Words Associated w/ Health & 44 \\ 
& & Words Associated w/ Uncertainty & 0\\ 
& & Simply Claiming Authority or Credibility & 234 \\\cline{2-4} 
& Call to Action & Total & 132 \\ 
\hline

\end{tabular}
\caption{Exhaustive Table of all Parent, Child, and Sub Annotations, along with the number of annotated examples for each.}
\label{child-sub}
\end{centering}
\end{table*}
\section{Label Normalization}
\label{appendix:label-norm}

As mentioned in Section \ref{subsection:models} our \textit{Health subset} labels are highly imbalanced. Since our data is heterogeneous and is gathered from multiple domains, it contains various similar labels. We normalize our labels using Table \ref{NormLabels} to normalize similar labels such as $false$ and $mostly false$ into the same category.

\begin{table*}[h!]
\begin{centering}
\begin{tabular}{ |l|l| } 

\hline
\textbf{Norm. Label} & \textbf{Original Label} \\
\hline
True & mostly true, truth!, true, in-the-green, mostly truth!\\
\hline
False & mostly false, fiction!, false \\
\hline

\end{tabular}
\caption{Table of all labels and our normalized versions of them}
\label{NormLabels}
\end{centering}
\end{table*}

\section{Result of Persuasive Strategy Detection}
\label{appendix:result}
The Precision, Recall, and F1-score of our best-performing model, along with the number of instances for each label, are presented for levels 1 to 4 in the following tables: level 1 in Tables~\ref{layer1results_no}, \ref{layer1results_low}, and \ref{layer1results_high}; level 2 in Tables~\ref{layer2results_no}, \ref{layer2results_low}, and \ref{layer2results_high}; level 3 in Tables~\ref{layer3results_none}, \ref{layer3results_low}, and \ref{layer3results_high}; and level 4 in Tables~\ref{layer4results_no}, \ref{layer4results_low}, and \ref{layer4results_high}, respectively.

\begin{table*}[h!]
\begin{centering}
\begin{tabular}{ |l|l|l|l|l| } 
\hline
\multicolumn{5}{|c|}{\textbf{No Context}} \\
\hline
& \textbf{Precision} & \textbf{Recall} & \textbf{F1} & \textbf{\# of Instances} \\
\hline
No Persuasive Writing Strategy & 0.647 & 0.586 & 0.615 & 633\\
Has Persuasive Writing Strategy & 0.690 & 0.742 & 0.715 & 784\\
\hline
\end{tabular}

\caption{Precision, Recall, F1 Score, and Support for the Test Set Level 1 of the persuasive strategy label's hierarchy with no context.}
\label{layer1results_no}
\end{centering}
\end{table*}

\begin{table*}[h!]
\begin{centering}
\begin{tabular}{ |l|l|l|l|l| } 
\hline
\multicolumn{5}{|c|}{\textbf{Low Context}} \\
\hline
& \textbf{Precision} & \textbf{Recall} & \textbf{F1} & \textbf{\# of Instances} \\
\hline
No Persuasive Writing Strategy & 0.657 & 0.670 & 0.664 & 633\\
Has Persuasive Writing Strategy & 0.729 & 0.718 & 0.724 & 784\\
\hline
\end{tabular}
\caption{Precision, Recall, F1 Score, and Support for the Test Set Level 1 of the persuasive strategy label's hierarchy with low context.}
\label{layer1results_low}
\end{centering}
\end{table*}

\begin{table*}[h!]
\begin{centering}
\begin{tabular}{ |l|l|l|l|l| } 
\hline
\multicolumn{5}{|c|}{\textbf{High Context}} \\
\hline
& \textbf{Precision} & \textbf{Recall} & \textbf{F1} & \textbf{\# of Instances} \\
\hline
No Persuasive Writing Strategy & 0.679 & 0.660 & 0.669 & 633\\
Has Persuasive Writing Strategy & 0.732 & 0.747 & 0.739 & 784\\
\hline
\end{tabular}
\caption{Precision, Recall, F1 Score, and Support for the Test Set Level 1 of the persuasive strategy label's hierarchy with high context.}
\label{layer1results_high}
\end{centering}
\end{table*}


\begin{table*}[h!]
\begin{centering}
\begin{tabular}{ |l|l|l|l|l| } 
\hline
\multicolumn{5}{|c|}{\textbf{No Context}} \\
\hline
& \textbf{Precision} & \textbf{Recall} & \textbf{F1} & \textbf{\# of Instances} \\
\hline
Narrative w/ Details & 0.629 & 0.629 & 0.629 & 105\\
Emotional Appeals & 0.352 & 0.292 & 0.319 & 65\\
Distinctive Linguistic Features & 0.537 & 0.797 & 0.642 & 182\\
Establishing Legitimacy & 0.585 & 0.414 & 0.485 & 266\\
Using Anecdotes and Personal Exp. as Evidence & 0.795 & 0.318 & 0.455 & 110\\
Distrusting Govt. or Corps. & 0.583 & 0.356 & 0.442 & 59\\
Politicizing Health Issues & 0.000 & 0.000 & 0.000 & 17\\
Highlighting Uncertainty and Risk & 0.300 & 0.240 & 0.267 & 25\\
Inappropriate Use of Scientific Evidence & 0.316 & 0.154 & 0.207 & 78\\
Rhetorical Tricks & 0.273 & 0.240 & 0.255 & 25\\
Biased Reasoning to Make Conclusion & 0.164 & 0.123 & 0.141 & 73\\
\hline
\hline
\end{tabular}
\caption{Precision, Recall, F1 Score, and Support for the Test Set Level 2 of the persuasive strategy label's hierarchy with no context.}
\label{layer2results_no}
\end{centering}
\end{table*}

\begin{table*}[h!]
\begin{centering}
\begin{tabular}{ |l|l|l|l|l| } 
\hline
\multicolumn{5}{|c|}{\textbf{Low Context}} \\
\hline
& \textbf{Precision} & \textbf{Recall} & \textbf{F1} & \textbf{\# of Instances} \\
\hline
Narrative w/ Details & 0.607 & 0.705 & 0.652 & 105 \\
Emotional Appeals & 0.250 & 0.231 & 0.240 & 65  \\
Distinctive Linguistic Features & 0.621 & 0.720 & 0.667 & 182 \\
Establishing Legitimacy & 0.572 & 0.523 & 0.546 & 266 \\
Using Anecdotes and Personal Exp. as Evidence & 0.745 & 0.373 & 0.497 & 110 \\
Distrusting Govt. or Corps. & 0.400 & 0.373 & 0.386 & 59  \\
Politicizing Health Issues & 0.154 & 0.118 & 0.133 & 17  \\
Highlighting Uncertainty and Risk & 0.474 & 0.360 & 0.409 & 25  \\
Inappropriate Use of Scientific Evidence & 0.338 & 0.333 & 0.335 & 78  \\
Rhetorical Tricks & 0.119 & 0.200 & 0.149 & 25  \\
Biased Reasoning to Make Conclusion & 0.260 & 0.178 & 0.211 & 73  \\
\hline
\end{tabular}
\caption{Precision, Recall, F1 Score, and Support for the Test Set Level 2 of the persuasive strategy label's hierarchy with low context.}
\label{layer2results_low}
\end{centering}
\end{table*}

\begin{table*}[h!]
\begin{centering}
\begin{tabular}{ |l|l|l|l|l| } 
\hline
\multicolumn{5}{|c|}{\textbf{High Context}} \\
\hline
& \textbf{Precision} & \textbf{Recall} & \textbf{F1} & \textbf{\# of Instances} \\
\hline
Narrative w/ Details & 0.664 & 0.676 & 0.670 & 105 \\
Emotional Appeals & 0.261 & 0.185 & 0.216 & 65 \\
Distinctive Linguistic Features & 0.586 & 0.863 & 0.698 & 182 \\
Establishing Legitimacy & 0.554 & 0.560 & 0.557 & 266 \\
Using Anecdotes and Personal Exp. as Evidence & 0.733 & 0.300 & 0.426 & 110 \\
Distrusting Govt. or Corps. & 0.309 & 0.356 & 0.331 & 59 \\
Politicizing Health Issues & 0.364 & 0.235 & 0.286 & 17 \\
Highlighting Uncertainty and Risk & 0.526 & 0.400 & 0.455 & 25 \\
Inappropriate Use of Scientific Evidence & 0.429 & 0.308 & 0.358 & 78 \\
Rhetorical Tricks & 0.125 & 0.040 & 0.061 & 25 \\
Biased Reasoning to Make Conclusion & 0.260 & 0.178 & 0.211 & 73 \\

\hline
\end{tabular}
\caption{Precision, Recall, F1 Score, and Support for the Test Set Level 2 of the persuasive strategy label's hierarchy with high context.}
\label{layer2results_high}
\end{centering}
\end{table*}


\begin{table*}[h!]
\begin{centering}
\begin{tabular}{ |l|l|l|l|l| } 
\hline
\multicolumn{5}{|c|}{\textbf{No Context}} \\
\hline
& \textbf{Precision} & \textbf{Recall} & \textbf{F1} & \textbf{\# of Instances} \\
\hline
Fear & 0.091 & 0.042 & 0.057 & 24 \\
Anger & 0.000 & 0.000 & 0.000 & 25 \\
Hope & 0.000 & 0.000 & 0.000 & 4 \\
Anxiety & 0.000 & 0.000 & 0.000 & 7 \\
Uppercase Words & 0.824 & 0.864 & 0.843 & 81 \\
Title of Article as Clickbait & 0.286 & 0.750 & 0.414 & 8 \\
Bolded, Underline or Italicized & 0.480 & 0.758 & 0.587 & 62 \\
Exaggerated Usage of Punctuation Marks & 0.581 & 0.783 & 0.667 & 46 \\
Citing Source to Establish Legitimacy & 0.459 & 0.487 & 0.473 & 80 \\
Legitimate Persuasive Techniques & 0.618 & 0.596 & 0.607 & 57 \\
Surface Credibility Markers & 0.519 & 0.456 & 0.485 & 90 \\
Call to Action & 0.486 & 0.375 & 0.424 & 48 \\
Verified to be False & 0.240 & 0.222 & 0.231 & 27 \\
Details Verified to be True & 0.607 & 0.270 & 0.374 & 63 \\
Details Not Verified & 0.200 & 0.250 & 0.222 & 16 \\
Financial Motivation & 0.400 & 0.205 & 0.271 & 39 \\
Out of Context\_Verified & 0.000 & 0.000 & 0.000 & 34 \\
Verified & 0.286 & 0.167 & 0.211 & 12 \\
Exaggeration/Absolute Language & 0.500 & 0.105 & 0.174 & 19 \\
Inappropriate Analogy or False Connection & 0.167 & 0.100 & 0.125 & 10 \\
Claims Without Evidence & 0.233 & 0.194 & 0.212 & 36 \\
Evidence Does Not Support Conclusion & 0.000 & 0.000 & 0.000 & 10 \\
Shifting Hypothesis & 0.000 & 0.000 & 0.000 & 14 \\

\hline
\end{tabular}
\caption{Precision, Recall, F1 Score, and Support for the Test Set Level 3 of the persuasive strategy label's hierarchy with no context.}
\label{layer3results_none}
\end{centering}
\end{table*}

\begin{table*}[h!]
\begin{centering}
\begin{tabular}{ |l|l|l|l|l| } 
\hline
\multicolumn{5}{|c|}{\textbf{Low Context}} \\
\hline
& \textbf{Precision} & \textbf{Recall} & \textbf{F1} & \textbf{\# of Instances} \\
\hline

Fear & 0.167 & 0.042 & 0.067 & 24 \\
Anger & 0.000 & 0.000 & 0.000 & 25 \\
Hope & 0.000 & 0.000 & 0.000 & 4 \\
Anxiety & 1.000 & 0.143 & 0.250 & 7 \\
Uppercase Words & 0.697 & 0.852 & 0.767 & 81 \\
Title of Article as Clickbait & 0.286 & 0.750 & 0.414 & 8 \\
Bolded, Underline or Italicized & 0.586 & 0.823 & 0.685 & 62 \\
Exaggerated Usage of Punctuation Marks & 0.635 & 0.717 & 0.673 & 46 \\
Citing Source to Establish Legitimacy & 0.427 & 0.512 & 0.466 & 80 \\
Legitimate Persuasive Techniques & 0.551 & 0.667 & 0.603 & 57 \\
Surface Credibility Markers & 0.528 & 0.311 & 0.392 & 90 \\
Call to Action & 0.655 & 0.396 & 0.494 & 48 \\
Verified to be False & 0.550 & 0.407 & 0.468 & 27 \\
Details Verified to be True & 0.852 & 0.365 & 0.511 & 63 \\
Details Not Verified & 0.222 & 0.125 & 0.160 & 16 \\
Financial Motivation & 0.400 & 0.103 & 0.163 & 39 \\
Out of Context\_Verified & 0.200 & 0.029 & 0.051 & 34 \\
Verified & 0.000 & 0.000 & 0.000 & 12 \\
Exaggeration/Absolute Language & 0.000 & 0.000 & 0.000 & 19 \\
Inappropriate Analogy or False Connection & 0.000 & 0.000 & 0.000 & 10 \\
Claims Without Evidence & 0.346 & 0.250 & 0.290 & 36 \\
Evidence Does Not Support Conclusion & 0.000 & 0.000 & 0.000 & 10 \\
Shifting Hypothesis & 0.125 & 0.071 & 0.091 & 14 \\
\hline
\end{tabular}
\caption{Precision, Recall, F1 Score, and Support for the Test Set Level 3 of the persuasive strategy label's hierarchy with low context.}
\label{layer3results_low}
\end{centering}
\end{table*}

\medskip
\medskip

\begin{table*}[h!]
\begin{centering}
\begin{tabular}{ |l|l|l|l|l| } 
\hline
\multicolumn{5}{|c|}{\textbf{High Context}} \\
\hline
& \textbf{Precision} & \textbf{Recall} & \textbf{F1} & \textbf{\# of Instances} \\
\hline
Fear & 0.211 & 0.167 & 0.186 & 24 \\
Anger & 0.000 & 0.000 & 0.000 & 25 \\
Hope & 0.000 & 0.000 & 0.000 & 4 \\
Anxiety & 0.167 & 0.143 & 0.154 & 7 \\
Uppercase Words & 0.724 & 0.877 & 0.793 & 81 \\
Title of Article as Clickbait & 0.222 & 0.500 & 0.308 & 8 \\
Bolded, Underline or Italicized & 0.621 & 0.871 & 0.725 & 62 \\
Exaggerated Usage of Punctuation Marks & 0.621 & 0.783 & 0.692 & 46 \\
Citing Source to Establish Legitimacy & 0.519 & 0.500 & 0.510 & 80 \\
Legitimate Persuasive Techniques & 0.603 & 0.667 & 0.633 & 57 \\
Surface Credibility Markers & 0.527 & 0.433 & 0.476 & 90 \\
Call to Action & 0.485 & 0.333 & 0.395 & 48 \\
Verified to be False & 0.237 & 0.667 & 0.350 & 27 \\
Details Verified to be True & 0.704 & 0.302 & 0.422 & 63 \\
Details Not Verified & 0.636 & 0.438 & 0.519 & 16 \\
Financial Motivation & 0.526 & 0.256 & 0.345 & 39 \\
Out of Context\_Verified & 0.000 & 0.000 & 0.000 & 34 \\
Verified & 0.000 & 0.000 & 0.000 & 12 \\
Exaggeration/Absolute Language & 0.125 & 0.053 & 0.074 & 19 \\
Inappropriate Analogy or False Connection & 0.000 & 0.000 & 0.000 & 10 \\
Claims Without Evidence & 0.417 & 0.278 & 0.333 & 36 \\
Evidence Does Not Support Conclusion & 0.000 & 0.000 & 0.000 & 10 \\
Shifting Hypothesis & 0.000 & 0.000 & 0.000 & 14 \\
\hline
\end{tabular}
\caption{Precision, Recall, F1 Score, and Support for the Test Set Level 3 of the persuasive strategy label's hierarchy with high context.}
\label{layer3results_high}
\end{centering}
\end{table*}


\begin{table*}[h!]
\begin{centering}
\begin{tabular}{ |l|l|l|l|l| } 
\hline
\multicolumn{5}{|c|}{\textbf{No Context}} \\
\hline
& \textbf{Precision} & \textbf{Recall} & \textbf{F1} & \textbf{\# of Instances} \\
\hline
source verified to be credible & 0.250 & 0.067 & 0.105 & 30 \\
source verified to not be credible & 0.000 & 0.000 & 0.000 & 6 \\
source not verified & 0.389 & 0.477 & 0.429 & 44 \\
rhetorical question & 0.674 & 0.574 & 0.620 & 54 \\
medical or scientific jargon & 0.400 & 0.167 & 0.235 & 12 \\
words associated with nature or healthiness & 0.400 & 0.167 & 0.235 & 12 \\
simply claiming authority or credibility & 0.596 & 0.456 & 0.517 & 68 \\

\hline
\end{tabular}
\caption{Precision, Recall, F1 Score, and Support for the Test Set- Level 4 of the persuasive strategy label's hierarchy with no context.}
\label{layer4results_no}
\end{centering}
\end{table*}

\begin{table*}[h!]
\begin{centering}
\begin{tabular}{ |l|l|l|l|l| }
\hline
\multicolumn{5}{|c|}{\textbf{Low Context}} \\
\hline
& \textbf{Precision} & \textbf{Recall} & \textbf{F1} & \textbf{\# of Instances} \\
\hline
source verified to be credible    & 0.375 & 0.100 & 0.158 & 30 \\
source verified to not be credible      & 0.000 & 0.000 & 0.000 &  6 \\
source not verified     & 0.412 & 0.477 & 0.442 & 44 \\
rhetorical question     & 0.544 & 0.685 & 0.607 & 54 \\
medical or scientific jargon   & 1.000 & 0.250 & 0.400 & 12 \\
words associated with nature or healthiness   & 0.000 & 0.000 & 0.000 & 12 \\
simply claiming authority or credibility     & 0.647 & 0.324 & 0.431 & 68 \\
\hline
\end{tabular}
\caption{Precision, Recall, F1 Score, and Support for the Test Set- Level 4 of the persuasive strategy label's hierarchy with low context.}
\label{layer4results_low}
\end{centering}
\end{table*}

\begin{table*}[h!]
\begin{centering}
\begin{tabular}{ |l|l|l|l|l| } 
\hline
\multicolumn{5}{|c|}{\textbf{High Context}} \\
\hline
& \textbf{Precision} & \textbf{Recall} & \textbf{F1} & \textbf{\# of Instances} \\
\hline
source verified to be credible    & 0.000 & 0.000 & 0.000 & 30 \\
source verified to not be credible      & 0.000 & 0.000 & 0.000 &  6 \\
source not verified     & 0.383 & 0.409 & 0.396 & 44 \\
rhetorical question     & 0.597 & 0.796 & 0.683 & 54 \\
medical or scientific jargon   & 1.000 & 0.167 & 0.286 & 12 \\
words associated with nature or healthiness   & 0.000 & 0.000 & 0.000 & 12 \\
simply claiming authority or credibility     & 0.667 & 0.324 & 0.436 & 68 \\
\hline
\end{tabular}
\caption{Precision, Recall, F1 Score, and Support for the Test Set- Level 4 of the persuasive strategy label's hierarchy with high context.}
\label{layer4results_high}
\end{centering}
\end{table*}

\end{document}